\documentclass{article}

\usepackage{PRIMEarxiv}

\usepackage[utf8]{inputenc} 
\usepackage[T1]{fontenc}    
\usepackage{hyperref}       
\usepackage{url}            
\usepackage{booktabs}       
\usepackage{amsfonts}       
\usepackage{amsmath}
\usepackage{amssymb}
\usepackage{pifont}
\usepackage{xcolor}
\usepackage{nicefrac}       
\usepackage{microtype}      
\usepackage{lipsum}
\usepackage{fancyhdr}       
\usepackage{graphicx}       
\usepackage{cite}

\newlength{\wholefigwidth}
\newlength{\smallfigwidth}
\newlength{\halfsmallfigwidth}
\newlength{\figwidth}
\newcommand{\Loss}{\mathcal{L}}
\newcommand{\Weights}{\boldsymbol\theta}
\newcommand{\Perturbation}{\boldsymbol\delta}

\definecolor{darkgreen}{RGB}{0, 150, 0}
\newcommand{\TrueMarker}{\textcolor{darkgreen}{\ding{51}}}
\newcommand{\FalseMarker}{\textcolor{red}{\ding{55}}}
  
\title{Data efficiency and extrapolation trends in neural network interatomic potentials}

\author{
  Joshua A. Vita\\
  University of Illinois at Urbana-Champaign, \\
  Lawrence Livermore National Laboratory \\
  \And   
  Daniel Schwalbe-Koda\thanks{Correspondence to: \url{dskoda@llnl.gov}}\\
  Lawrence Livermore National Laboratory
}

\begin{document}
\maketitle

\begin{abstract}
Over the last few years, key architectural advances have been proposed for neural network interatomic potentials (NNIPs), such as incorporating message-passing networks, equivariance, or many-body expansion terms. 
Although modern NNIP models exhibit small differences in energy/forces errors, improvements in accuracy are still considered the main target when developing new NNIP architectures.
In this work, we show how architectural and optimization choices influence the generalization of NNIPs, revealing trends in molecular dynamics (MD) stability, data efficiency, and loss landscapes.
Using the 3BPA dataset, we show that test errors in NNIP follow a scaling relation and can be robust to noise, but cannot predict MD stability in the high-accuracy regime.
To circumvent this problem, we propose the use of loss landscape visualizations and a metric of loss entropy for predicting the generalization power of NNIPs.
With a large-scale study on NequIP and MACE, we show that the loss entropy predicts out-of-distribution error and MD stability despite being computed only on the training set.
Using this probe, we demonstrate how the choice of optimizers, loss function weighting, data normalization, and other architectural decisions influence the extrapolation behavior of NNIPs.
Finally, we relate loss entropy to data efficiency, demonstrating that flatter landscapes also predict learning curve slopes.
Our work provides a deep learning justification for the extrapolation performance of many common NNIPs, and introduces tools beyond accuracy metrics that can be used to inform the development of next-generation models.
\end{abstract}

\section{Introduction}

Machine learning (ML) has proven extremely valuable in the materials and chemical sciences as a tool for data analysis and generation \cite{Bartok2017,Butler2018,Schmidt2019,Keith2021Combining}.
Particularly in atomistic simulations, ML-based models offer a compelling balance between high-accuracy, high-cost quantum chemistry calculations and low-accuracy, low-cost classical force fields \cite{Behler2015,Mueller2020Machine,Manzhos2020a}.
Whereas several models based on kernel regression or Gaussian processes have been proposed \cite{Mueller2020Machine,Chmiela2017,Unke2021a,Bartok2010,Christensen2020fchl,Vandermause2020,Deringer2021Gaussian}, recent developments in neural network (NN) interatomic potentials (IPs) have shown promise due to their low inference time, scalability to large datasets, and high accuracy in predicting potential energy surfaces (PESes) \cite{Manzhos2020a,Unke2021a,Behler2007}.
These methods have been used for a variety of applications, including molecular simulation, excited-state dynamics, phase transitions, chemical reactions, and more \cite{Unke2021a,Behler2008,Cheng2019,westermayr2022deep,manzhos2020neural}.

Despite their successes, NNIPs still struggle with data efficiency and robust generalization.
Over the last few years, several different model architectures were proposed to reduce errors in PES fitting, decrease the amount of data required to train the models, and improve predictions for configurations beyond the training domain.
In particular, NN architectures incorporating physics concepts such as directional representations and equivariance \cite{Kondor2018,Thomas2018,Anderson2019,Mailoa2019,Batzner2022} or many-body interactions \cite{Batatia2022a,Batatia2022, Musaelian2022} have gained popularity due to higher accuracy and data efficiency.
Nevertheless, recent works show that accuracy metrics over datasets are insufficient to quantify the models' quality in production simulations and motivate the use of alternative metrics such as computational speed or simulation stability \cite{Zuo2020,Kovacs2021,Fu2022,Stocker2022,Morrow2022validate,Vita2021}.
Different NNIP models may have similar test accuracy, but completely different extrapolation ability \cite{Wellawatte2022,Schwalbe-Koda2021}.
This begs the question: \textbf{which metrics can distinguish between NNIPs with similar test error but different extrapolation behavior?}

In this work, we investigate trends in robustness and extrapolation behavior of NNIPs, and propose a metric to predict their stability in production simulations and data efficiency.
In particular, we provide the following contributions:

\begin{itemize}
    \item Using literature data and two state-of-the-art NNIP models (NequIP and MACE), we show that extrapolation trends can be obtained from error metrics in the 3BPA dataset. For example, the NNIPs are able to recover the underlying PES despite being trained to noisy labels, and have extrapolation errors that follow scaling relations against in-domain accuracy metrics. Although these trends hold across different architectures, we show that this scaling relation is not recovered in the high-accuracy regime, and thus cannot be used to downselect robust models.    
    \item To circumvent the limitations above, we propose that loss landscapes (LLs) can provide evaluation strategies for NNIPs beyond accuracy metrics. Qualitative inspection of LLs explains some heuristic training regimes for NNIPs, such as the use of higher weights for forces loss, weight cycling, or separating learning rates for different parts of the architecture, thus providing theoretical justification for certain hyperparameters.
    \item Using a large-scale study with NequIP and MACE, we show that flatness of LLs correlates with model robustness. In particular, we quantify the loss entropy around the optimized models and relate them to errors in the extrapolation regime. Furthermore, we show that molecular dynamics simulations of models with flatter LLs exhibit less unphysical behavior than their sharper counterparts.
\end{itemize}

This combination of theoretical justification, benchmarking, and a new metric can aid the development of newer NNIP architectures with higher accuracy, trainability, and robustness.

\section{Background}
\label{sec:background}

\textbf{NNIPs:} NN-based force fields have been first proposed using feedforward NNs and symmetry-based representations \cite{Behler2007}. In these systems, improvements are achieved by designing new representations to better capture the atomic environment \cite{Behler2011, Jose2012, Smith2017, Huan2017universal,Zhang2018deep,Wood2018, Drautz2019}. More recently, message-passing neural networks (MPNNs) \cite{Gilmer2017} showed remarkable ability to fit PESes using learned representations. In this area, most works compare models according to their accuracy with respect to standardized datasets, such as QM9 \cite{Ruddigkeit2012,Ramakrishnan2014}, MD17 \cite{Chmiela2017}, and others. MPNN-based potentials often vary according to their interaction blocks, handling of symmetry operations, and general architectural choices.

\textbf{NN representation capacity and generalization:} although NNs have a large number of parameters, obtaining NN-based models with low generalization error is not uncommon. Preventing overfitting may require regularization techniques such as changing the loss function, e.g. with weight regularization or decay, augmenting the dataset, or using training protocols such as adaptive learning rates, dropout, and more \cite{Song2020,Keskar2016,Neyshabur2017}. However, these are not requirements to controlling the generalization error \cite{Zhang2016}. For example, NNs with good generalization capacity have been shown to overfit to random labels and input data \cite{Zhang2016}. This suggests that, given enough parameters, even architecturally regularized NNs exhibit wide representation capacity for arbitrary data sets. In some cases, however, training to noisy labels can only be overcome by adapting architectures, regularization techniques, and correcting the loss function \cite{Song2020}.

\textbf{Loss landscapes (LLs):} the shape of the LL is strongly correlated with the trainability of NN architectures. The minimization of the empirical risk is easier when the (non-convex) LL is smoother, as it yields more predictive gradients \cite{Goodfellow2016,Gilmer2021}. Furthermore, LLs with several local minima exhibit lower trainability than their smoother counterparts \cite{Chaudhari2016,Neyshabur2017}. Some works proposed that flatter LLs are also related to lower NN generalization error \cite{hochreiter1994simplifying,hinton1993keeping,Chaudhari2016}. Although this relationship is often complicated by factors such as batch size \cite{Keskar2016}, normalization \cite{Santurkar2018}, or weight decay, the sharpness of the LL is the most predictive of generalization error in NNs \cite{jiang2019fantastic}. Whereas this sharpness/flatness can be quantified using the Hessian of the loss \cite{Keskar2016,Neyshabur2017} or assumptions of a prior on the weights \cite{mcallester1999pac}, visualizations of the LL have proven useful to illustrate these minima with respect to the weight space \cite{Goodfellow2014,Li2017,Hao2019visualizing,Verpoort2020} without the cost associated to calculating the full Hessian. Analogies with potential energy landscapes have also been performed to explore the cost functions of other machine learning methods and explain the shape of these optimization landscapes \cite{ballard2017energy,verpoort2020archetypal}.

\section{Methods}

\subsection{Visualizing loss landscapes}
\label{sec:methods:ll}

The loss landscape $\ell$ of a neural network can be plotted by evaluating the loss function $\Loss$ along a trajectory between two parameter sets $\Weights$ and $\Weights'$.
The simplest approach is to linearly interpolate the weights \cite{Goodfellow2014}, choosing a scalar $t \in [0, 1]$ such that $\Weights (t) = (1 - t) \Weights + t \Weights'$. Then, the loss landscape $\ell$ for a model becomes

\begin{equation}\label{eq:ll:interpolation}
    \ell(t) = \Loss \left( \Weights (t) \right) = \Loss \left( (1 - t) \Weights + t \Weights' \right).
\end{equation}

\noindent In this work, the loss $\Loss$ is evaluated on the training set of the model with parameters $\Weights$. 

In the absence of the reference weights $\Weights'$, the LL can be constructed by sampling a random vector $\Perturbation$ in the parameter space and plotting the LL around $\Weights$ as

\begin{equation}\label{eq:ll:1D}
    \ell(t; \Perturbation) = \Loss \left( \Weights + t \Perturbation \right),
\end{equation}

\noindent where the domain of $t$ is appropriately chosen to span a neighborhood of $\Weights$. This notion can be extended to 2D LLs by taking two orthogonal vectors $\Perturbation_1, \Perturbation_2$ such that

\begin{equation}\label{eq:ll:2D}
    \ell(t_1, t_2; \Perturbation_1, \Perturbation_2) = \Loss \left( \Weights + t_1 \Perturbation_1 + t_2 \Perturbation_2 \right),
\end{equation}

\noindent with scalars $t_1, t_2$ chosen to span a (two-dimensional) neighborhood of $\Weights$. These approaches have been used to study the LLs of NN classifiers in image datasets, interpolate between sets of classifiers, and explore the loss function around degenerate minima \cite{Keskar2016, Im2016, Nguyen2017, Smith2017LL}.

One challenge when analyzing LLs is comparing different models according to their parameters. Because activation functions such as ReLU allow for scale-invariance of NN weights, especially when coupled with batch normalization techniques, the magnitude of the vector $\Perturbation$ is not transferable from model to model. This prevents a fair comparison of LLs, curvatures, and sharpness metrics. To account for this, we use the filter normalization technique proposed by Li et al. \cite{Li2017}. Therein, each random vector $\Perturbation$ is normalized by the scale of each filter $i$, in each layer $j$ of the NN, i.e.

\begin{equation}\label{eq:ll:norm}
    \bar{\Perturbation}^{(i,j)} = \frac{\Perturbation^{(i,j)}}{\lVert \Perturbation^{(i,j)} \rVert} \lVert \Weights^{(i,j)} \rVert,
\end{equation}

\noindent where $\lVert . \rVert$ is the Frobenius norm. Then, the LL is plotted according to the filter-normalized vector $\bar{\Perturbation}$,

\begin{equation}\label{eq:ll:norm-interpolation}
    \ell(t; \bar{\Perturbation}) = \Loss \left( \Weights + t \bar{\Perturbation} \right),
\end{equation}

\noindent and analogously for 2D LLs.

Although informative, sampling LLs can have cost comparable to training the NN, since evaluating the loss for each interpolated weight in each direction is equivalent to one training epoch.
Depending on the number of parameters and directions $\Perturbation_n$ under analysis, the loss may be evaluated over the entire training dataset a large number of times.

\subsection{Quantitative comparison of loss landscapes}\label{sec:methods:entropy}

Despite the usefulness of visualizing loss landscapes, comparing them beyond qualitative insights requires metrics to differentiate them. The most commonly used metric in the field is the curvature of the LL \cite{jiang2019fantastic}, which is related to the magnitude of the eigenvalues of the loss Hessian. Although informative, the Hessian is a local property and cannot fully capture ``valley-like'' degeneracies in loss landscapes. Furthermore, computing the full Hessian for a system with millions of training parameters would be intractable. Thus, alternative metrics to compare LLs are required.

In the case of NNIPs, which are often trained both to energies and forces, two LLs are obtained per model, one for each target value being trained to. Although they cannot be completely disentangled, both LLs should be compared simultaneously to assess model performance. To derive one of such metrics, we propose the use of ``loss entropy'' \cite{Chaudhari2016,baldassi2020shaping} to quantify loss flatness around optimized minima. 
Although variations of this quantity he been proposed, we use the formula

\begin{equation}\label{eq:ls:entropy}
    S(T) = k \log \left[
    \sum_{t} \exp \left(
    \frac{-\bar{\ell} \left( t \right)}{kT}
    \right)
    \right],
\end{equation}

\noindent where $S$ is the entropy of the loss landscape $\bar{\ell} (t)$ computed with respect to energy or forces and averaged over $N$ orthogonal, filter-normalized weight displacements,

\begin{equation}\label{eq:ls:entropy}
    \bar{\ell} \left( t \right) = 
    \frac{1}{N}
    \sum_n \ell (t; \bar{\Perturbation}_n).
\end{equation}

\noindent and $kT$ is a weighting parameter that quantifies the ``flatness'' of the LL with respect to a certain acceptable threshold of training loss. This is similar to distributions of microstates accessible in a given ``temperature,'' and ensures that low-loss states contribute to a much larger entropy than high-loss states.

As the entropy of the LL of a NNIP takes into account both energy and forces losses, we adopt a strategy similar to the one used during training of the NNs by balancing energy and forces losses with a weighted sum,

\begin{equation}\label{eq:ls:weighted_S}
    S = \alpha S_E(T_E) + (1 - \alpha) S_F (T_F),
\end{equation}

\noindent where $\alpha$ is a dimensionless parameter between 0 and 1 that weights the entropy of the energy loss ($S_E$) and forces loss ($S_F$). As these losses have different units, they have to be computed with different normalization parameters $k_E T_E$ and $k_F T_F$, each of which takes into account ``thermal randomness'' of the training loss. For simplicity, in this work, we adopt $k = 1$ as a dimensionless parameter, and assign the adequate units to the ``thermal error'' for better interpretability of the loss entropy.

\subsection{NNIPs and dataset}

\textbf{NequIP} \cite{Batzner2022} is an equivariant NNIP that uses Clebsch-Gordon transformations and spherical harmonics to incorporate equivariance in the model. NequIP demonstrates state-of-the-art accuracy in several datasets,  data efficiency, and has been employed to simulate a variety of organic and inorganic systems.

\textbf{MACE} \cite{Batatia2022} is an equivariant NNIP that uses higher-order messages to efficiently embed information beyond two-body interactions in traditional MPNNs. The model demonstrates state-of-the-art performance in a variety of benchmarks, faster learning, and competitive computational cost.

\textbf{Other NNIPs} proposed recently and not benchmarked in this study include Allegro \cite{Musaelian2022}, GemNet \cite{Gasteiger2021}, DimeNet \cite{Gasteiger2020}, HIP-NN \cite{Lubbers2018}, NewtonNet \cite{Haghighatlari2021}, BOTNet \cite{Batatia2022a}, SchNet \cite{Schutt2018}, PaiNN \cite{Schutt2021}, and others \cite{Smith2017,Zhang2018deep,Anderson2019,Hu2021forcenet}.

Training details are provided in Appendix \ref{sec:si:training}.

\textbf{The 3BPA dataset} \cite{Kovacs2021} under study in this work was chosen due to its previous use in the literature for benchmarking NNIP models in extrapolation behavior \cite{Batatia2022, Batatia2022a, Musaelian2022}.
The benchmark involves training models on low-temperature samples and evaluating their performance on held-out samples from high-temperature simulations.
Distributions of energies and forces of the 3BPA dataset are shown in Appendix \ref{sec:si:additional_3bpa}, Fig. \ref{fig:si:hist_3bpa}. Additional results using the ANI-Al dataset \cite{Smith2021} be found in Appendix \ref{sec:si:additional_ani}.

\section{Results and Discussion}

\subsection{Robustness to noise and extrapolation trends in NNIPs}
\label{sec:results:representation}

When comparing NNIPs, metrics of interest typically include errors in predicting forces and energies of a test dataset, and are appropriately used as baselines for assessing model quality.
However, accuracy metrics are not necessarily predictive of extrapolation power \cite{Fu2022}.
A first hypothesis to consider when analyzing NNIP extrapolation is whether NNIPs can learn a PES despite being trained on noisy data.
Although this test does not measure extrapolation to out-of-distribution data, it verifies whether models are expected to overfit to corrupted training data, which would lower their robust generalization power.
For example, NN-based classifiers can overfit to random labels in image datasets or to completely random inputs \cite{Zhang2016}, even in architectures with good generalization error that are designed to prevent overfitting.
Most NNIPs have enough parameters to memorize the training data, but standard regularization and architectural choices can curb overfitting in NNIPs, leading to lower generalization errors.

\begin{figure*}[!h]
    \centering
    \includegraphics[width=\linewidth]{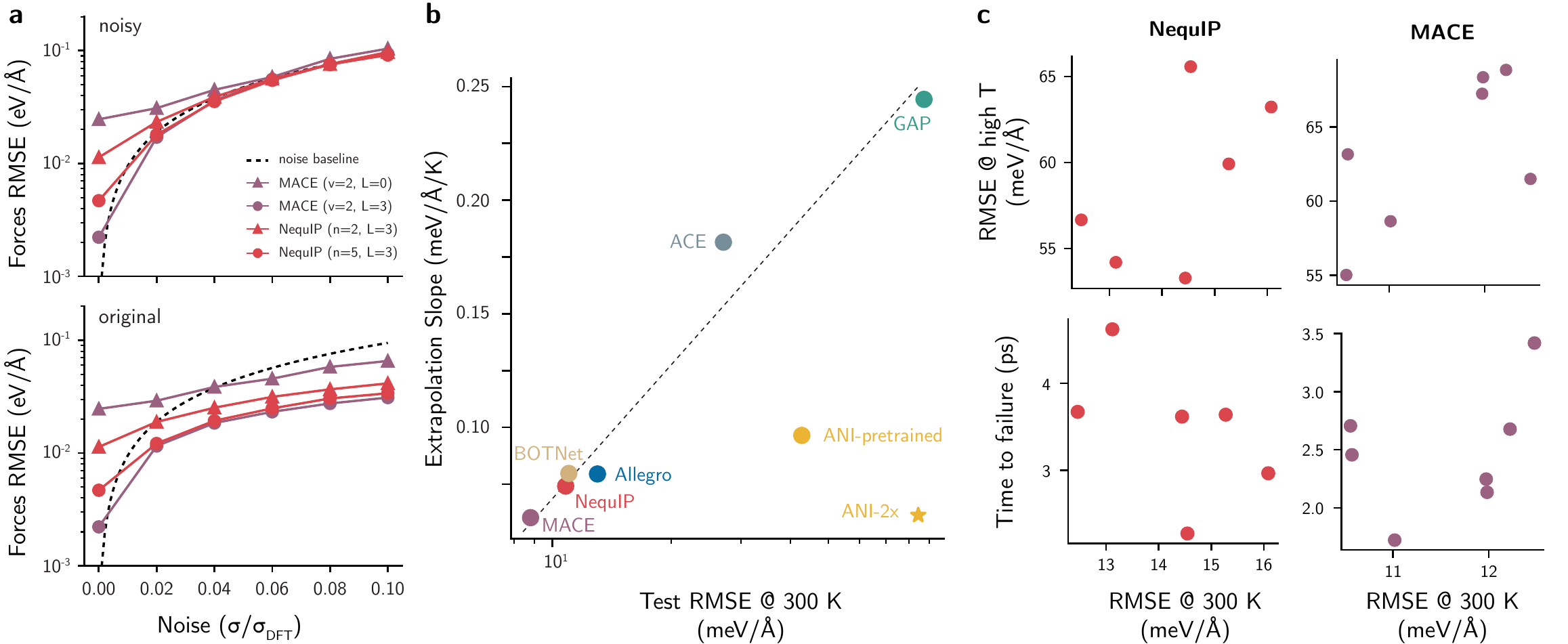}
    \caption{
    \textbf{a}, Force RMSE values (eV/\AA) for models trained to only the forces of increasingly noisy versions of the 3BPA dataset.
    The models trained on noisy data are then evaluated on their noisy training set (``noisy'' panel) or on the original, uncorrupted dataset (``original''). The  dashed lines correspond to the amount of noise that was added to the DFT forces in units of eV/\AA. The legend denotes the maximum body order of the MACE models ($v$), the number of interaction layers in the NequIP models ($n$), or the maximum order of the model irreps ($L$). All MACE models in \textbf{a} used only 2 interaction layers.
    \textbf{b}, Relationship between extrapolation ability, quantified using the slope of the test errors on the 3BPA dataset with increasing temperature (``extrapolation slope''), and the test error at 300 K. Slopes were computed by performing a linear fit to values taken from the literature for MACE \cite{Batatia2022}, BOTNet \cite{Batatia2022a}, and others \cite{Musaelian2022} (see Fig. \ref{fig:si:generalizability_fits}). The dashed line is a visual guide.
    \textbf{c}, Forces RMSE on the 300 K test set of several NequIP and MACE models with different architectures, but similar testing errors on the 300 K split. (top) Correlation between test errors on all splits (300, 600, and 1200 K) and test error on the 300 K split for models with similar test errors. (bottom) Average time-to-failure of MD simulation for each of the models with similar test errors.
    }
    \label{fig:noisy_fits}
\end{figure*}

To test this hypothesis, we trained four different NNIPs to the 3BPA dataset and analyzed their \textit{training} error.
Following the lead from the deep learning literature \cite{Zhang2016}, we then gradually corrupted the labels of the training set by randomly adding a sample from $\mathcal{N}(0, \sigma \cdot \sigma_{\text{DFT}})$ to the true forces, where $\sigma_{\text{DFT}}$ is the standard deviation in the forces predicted by density functional theory (DFT), and $\sigma$ is a scalar ranging from 0.0 to 0.1 (see Appendix \ref{sec:si:additional_3bpa}).
In principle, NN regressors with arbitrary levels of expressivity (or absent regularization) could achieve low training error even in these noisy PESes.
Figure \ref{fig:noisy_fits}a shows the error for NNIPs trained to the 3BPA dataset with corrupted forces, and tested on the original, un-corrupted data.
When the forces are not corrupted, models exhibit reasonable training errors lower than 40 meV/\AA, as expected by their nominal performances in energy prediction \cite{Batzner2022,Batatia2022}.
However, even small amounts of noise in the forces prevent the noisy dataset from being memorized with high accuracy, with the training loss plateauing instead of tending to zero.
The ability of the models to predict the noisy forces saturates at the limit of the noise, indicating that these NNIPs do not memorize the high-frequency labels.
On the other hand, when the test error of the models trained with corrupted labels is computed with respect to the uncorrupted dataset, the error is substantially smaller than the noise baseline (see the ``original'' panel of Fig. \ref{fig:noisy_fits}a).
Thus, the NNIPs under analysis are able to learn the underlying PES in the 3BPA dataset despite the added noise. 

Contrary to the overfitting hypothesis, these results suggest that data redundancy in the training set may help NNIPs to ``denoise'' the data.
To illustrate this effect, we show in Appendix \ref{sec:linregress} how data redundancy downplays the effect of external noise in a toy system.
Fig. \ref{fig:si:linregress_noise} shows how a large number of training points can counterbalance the effect of external noise when predicting the original, non-noisy data for the case of linear regressor models.
In this case, the model averages out the noise and recovers the true function even at high levels of added noise.
On the other hand, at the low-data regime, the regression model is unable to recover the true function, and its error quickly grows.
Although the results from the linear model may not directly translate to the case of NNs, Fig. \ref{fig:noisy_fits}a shows that, to an extent, NNIPs are able to ``denoise'' energies from the dataset due to architectural, training, or implicit data regularization.

To verify if this behavior was specific to the 3BPA dataset or could generalize to other datasets, we corrupted the energies or forces of the ANI-Al dataset and trained different models on the noisy values (see Appendix \ref{sec:si:additional_ani}), obtaining similar results even when only energies were considered.
As the ANI-Al example exhibited even better results than the 3BPA, we tested whether a drastic increase in the injected noise could still be denoised by the NNIPs under study. 
Following the results of Fig. \ref{fig:si:linregress_noise}, we trained the two NNIP architectures under study to PESes with energy noises up to twenty times higher than those in Fig. \ref{fig:noisy_fits} (see Fig. \ref{fig:si:ani_high_noise}), and up to twice the standard deviation of the original dataset distribution of the ANI-Al dataset.
Although the distribution of per-atom energies shows that the noisy PES is completely different than the original one (Fig. \ref{fig:si:ani_high_noise}b,c), all models succeeded in modeling the underlying PES below the error baseline (Fig. \ref{fig:si:ani_high_noise}).
As also illustrated by the toy example, the performance of the models degrades as extremely large amounts of noise are added.
Nonetheless, errors with respect to the non-noisy dataset are remarkably low considering the corruption baselines.

The toy example from Appendix \ref{sec:linregress} can be considered an upper bound of the ``dataset denoising'' ability, given that a functional form of the inputs is known and the model could, in principle, fit perfectly to the data. 
As the trends of NNIPs trained on the 3BPA or ANI-Al potential approach this behavior, it can be concluded that the two NNIP architectures are able to ``denoise'' the external noise added to the datasets, possibly due to data redundancy.
As generalization tests assume the model is being tested on unseen data, it is not clear whether the accuracy reflects the quality of the model predictions or simply their ability to reproduce local environments existing in the training data.
This is particularly important in the high-data regime, when the test dataset may be correlated to the train dataset in non-obvious ways.

To bypass this problem, alternative strategies were proposed to measure generalization power, including separating train-test splits according to sampling temperature, as is the case of the 3BPA dataset \cite{Kovacs2021}.
While testing a model on held-out samples from high-temperature simulations is typically considered an independent evaluation of its performance, we have found that extrapolation errors are correlated with low-temperature test errors across various model architectures.
This is performed by fitting a linear model to the errors on the 3BPA testing sets at 300, 600, and 1200 K from the literature \cite{Batatia2022, Batatia2022a, Musaelian2022} (see Fig. \ref{fig:si:generalizability_fits}), then using the slope of the fitted line as associated metric.
Fig. \ref{fig:noisy_fits}b shows that all the models that were trained only to 3BPA frames at 300 K follow an approximate linear scaling relation between the extrapolation slope and the log of the low-temperature errors.
This correlation between the low- and high-temperature data does not strictly preclude the use of the 3BPA dataset for assessing a fitted model's extrapolation abilities, as the extrapolation slope represents how much the accuracy of a given model degrades as the sampling temperature increases.
Nevertheless, it suggests that data and model regularization effects may be enforcing extrapolation trends in wide error ranges.
For example, a model like ACE \cite{Drautz2019,Batatia2022a} is known to provide functional forms that aid extrapolation beyond the training data \cite{Kovacs2021}. Despite this, the ACE model for 3BPA follows the same trends of more complex message-passing NNs.
In fact, the generalization slopes are correlated to their low-temperature error regardless of significant architectural differences.
This indicates that, for this benchmark, the root mean squared errors (RMSEs) at higher temperatures can be estimated from the test error at 300 K even without evaluating the model at the other test sets.

The exceptions to this rule are the two ANI models (``ANI-2x'' and ``ANI-pretrained''), which were pre-trained to the 8.9 million configurations from the ANI-2x dataset \cite{Devereux2020}. As seen in other fields of deep learning \cite{Abnar2022exploring}, the pre-trained models extrapolate better than all other models, though fine-tuning on 3BPA (``ANI-pretrained'') leads to slightly worse extrapolation slope. These results suggest that: (1) more diverse datasets may be required for assessing the extrapolation and generalization capacity of a model; and (2) pre-training on large datasets may be required to create universal NNIPs \cite{chen2022universal}, given that pre-trained ANI models were able to escape the scaling relation seen in Fig. \ref{fig:noisy_fits}b.

Although the scaling relation can estimate the extrapolation slope within reasonable bounds, it is unable to recover trends within the same architecture in the low-error regime.
As best-performing models often have differences of force errors smaller than 5 meV/\AA~ in the 300 K test set, it is not clear whether their extrapolation behavior can be accurately recovered from the scaling relation.
Indeed, Fig. \ref{fig:noisy_fits}c shows the correlation between the forces RMSE computed on the 300 K split and all splits (300, 600, and 1200 K, ``high T'') for several NequIP and MACE models with different hyperparameters and training methods (see Tables \ref{tab:si:nequip_rmse} and \ref{tab:si:mace_rmse}), but similar testing error at the 300 K split.
Although there is a positive correlation between the two RMSE metrics, the dispersion of points indicates that models with nearly the same RMSE for the 300 K data split show discrepancies in error above 10 meV/\AA~ when all splits are taken into account.

This scenario is aggravated when molecular dynamics (MD) simulations are used to test the extrapolation power of the NNIPs. 
When measuring the average simulation times for each of the models (see Appendix \ref{sec:si:md} for details on the MD simulations), no clear correlation is obtained between the error on the 300 K split of 3BPA and the average (physically meaningful) simulation length (Fig. \ref{fig:noisy_fits}c).
This motivates the creation of a metric that captures robustness trends in NNIPs.

\subsection{Training insights derived from loss landscapes}
\label{sec:results:ll}

Beyond data regularization, architectural and training choices strongly influence generalization ability of NN models.
Relevant aspects include model initialization, hyperparameter optimization, choice of optimizer, batch sizes, and many others.
Although the process of training NNs strongly affects their extrapolation behavior, good NN models systematically outperform their counterparts by optimizing to better local minima in the LL \cite{Santurkar2018}.
Thus, using these insights, we propose that \textit{loss landscapes of NNIP models can predict their generalization ability towards unseen data despite using only the training data}.
Correlations between robust generalization and loss sharpness have been observed for other NN models in the literature \cite{Chaudhari2016,jiang2019fantastic}, but not yet explored in the context of NNIPs.

\begin{figure}[!h]
    \centering
    \includegraphics[width=\linewidth]{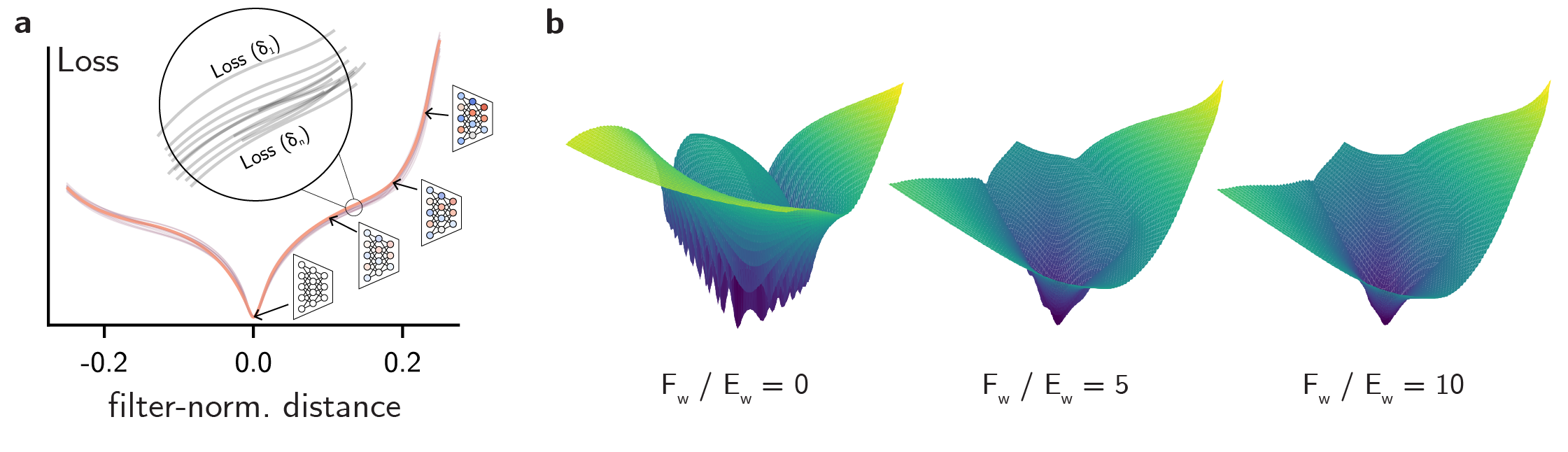}
    \caption{\textbf{a}, A schematic describing the process of generating loss landscapes. Starting at the origin with a trained model, landscapes are constructed by performing a grid-sampling along a random filter-normalized direction \cite{Li2017} in parameter space up to a chosen maximum distance. In order to ensure that that results were consistent regardless of the choice of random sampling direction, landscapes were averaged over multiple random directions, $\delta_1 \ldots \delta_n$, with $n=20$ usually showing only slight variations in landscape topography. \textbf{b}, 2D LLs for a MACE model trained to the 3BPA dataset, generated by sampling a plane defined by two random directions. The model was trained using $E_w=1$ and $F_w=1000.0$, but the final landscape was re-weighted for the purpose of this figure using a fixed value of $E_w = 1.0$ and $F_w$ increasing from 0.0 to 10.0 to demonstrate the effects of different weights on optimization.}
    \label{fig:ll_interpolation}
\end{figure}

To first verify if qualitative insights could be derived from LLs in the context of NNIPs, we investigated the behavior of the loss function around the optimized minima of the NNIP models trained to the non-noisy 3BPA dataset. 
To ensure the LL visualizations were statistically meaningful, we sampled 20 different orthogonal directions for each set of parameters and models, and interpolated them using the filter-normalized method described in Sec. \ref{sec:methods:ll} (see also Fig. \ref{fig:ll_interpolation}a). Then, we compared the LLs of the NNIP models using 2D visualizations, as often done for NN classifiers \cite{Li2017}.
Qualitative inspection of the 2D LLs in Fig. \ref{fig:ll_interpolation}b reveals the presence of weight degeneracies in the prediction of energies in models (see also Fig. \ref{fig:si:2d_ll_aluminum} for another example in the ANI-Al dataset). This ``valley-like'' landscape represents a subspace of weights leading to similar accuracy in energy \cite{draxler2018essentially}, and reflects the interplay between energy and forces during training.
These results agree with the literature regarding LLs of over-parameterized models \cite{Liu2022}, as well as the notion that physical systems often result in so-called ``sloppy'' models \cite{Gutenkunst2007, Kurniawan2022}, and can improve trainability and interpolation \cite{Bubeck2021}.

Qualitative analysis of LLs also explain other factors typically found as heuristics of NNIP training.
For example, the energy/forces coefficients in the final loss are often defined from hyperparameter optimization \cite{Schutt2017, Batzner2022} and have ratios varying from 1:10 to 1:100 for energy:forces RMSE.
Nevertheless, the success of these higher ratios can be justified from the perspective of the LL.
In Fig. \ref{fig:ll_interpolation}b, we show how a higher weight on force losses leads to LLs with less saddle points around the optimized minima, thus favoring training.
Although energy and forces are related and completely disentangling their effects is not achievable, the interpolation in Fig. \ref{fig:ll_interpolation}b shows how mixing both can lead to better optimization landscapes for NNIPs.
Based on these results, an effective training regimen would start with a relatively large weight for forces loss for a fast (and smoother) optimization of forces.
Once the force errors are reasonably converged, the weight can be decreased until a desired threshold in energy errors is achieved.
Similar strategies with weight cycling and scheduling --- e.g., starting with an energy:force loss weighting of 1:10, but then switch to 1000:1 in the later stages of training --- can also be effective.

One limitation of visualizations of the high-dimensional LL is that different directions in weight perturbation may lead to similar losses. This results in only slight variations in landscape topography with different random directions, as noted in Fig. \ref{fig:ll_interpolation}.
This may be related to the dominance of specific layers of the model in the filter normalization technique.
Figure \ref{fig:si:model_parameters} shows how the distribution of weights is non-uniform, suggesting some layers have higher sensitivity to weight perturbation than others.
As the filter normalization displaces the model weights along random directions with magnitude proportional to the norm of each filter and layer, parameters with higher weights may influence certain regions of the loss landscapes.
An exception to this point is when the parameters are intertwined with functions embedded in the architecture, such as the trainable Bessel functions in NequIP.
As shown in Fig. \ref{fig:si:ray_tracing_frozen}, freezing certain high-magnitude weights when generating the LLs can help flatten the landscape and remove spurious minima, emphasizing the importance of proper regularization and training regimen taking these effects into account (e.g., separate learning rates for certain layers in the model).

\subsection{Loss landscapes predict extrapolation trends in NequIP}
\label{sec:results:nequip}

In the context of NNIPs, robust generalization has important ramifications in two distinct areas: the stability of the model in production (e.g., MD simulations), and its data efficiency during training.
To probe the first of these features, we trained NequIP models using various choices of model architecture and optimization techniques (see Table \ref{tab:si:nequip_hparams}) in the low-temperature split of the 3BPA dataset.
Then, we performed MD simulations in the NVT ensemble with a high temperature of 1600 K (Appendix \ref{sec:si:md}) to ensure all models were extrapolating well beyond their training split.
Furthermore, as the 3BPA benchmark already provides geometries sampled from simulations up to 1200 K, the MD trajectories at 1600 K could provide information beyond the 1200 K splits provided by the dataset.
For each model in the study, 30 MD trajectories were simulated to obtain reasonable statistics on the model behavior, with simulation timelengths of up to 6 ps and a timestep of 1 fs.
In principle, a model's ability to extrapolate beyond the training set can be assessed by the number of trajectories that behave in a physical way \cite{Schwalbe-Koda2021}, as well as the average trajectory length until the model fails to extrapolate (Appendix \ref{sec:si:md}).

Using the results from MD simulations, we investigate whether LLs can predict the extrapolation behavior and trajectory stability of models.
While the test performance at low-temperature samples is unable to perform such predictions (Sec. \ref{sec:results:representation}), the test error at high-temperature samples is still expected to be a reasonable predictor of such stability.
Nevertheless, whereas high-temperature samples are available in the 3BPA dataset, out-of-domain data is often not available for an arbitrary material system.
In addition to being expensive to generate using ground-truth methods, sampling new configurations is rarely performed in an exhaustive way.
Thus, the major advantage of extrapolation tests based on LLs is their reliance only on the training dataset.
To quantify the sharpness of the LLs \cite{jiang2019fantastic}, we compute the loss entropy as described in Sec. \ref{sec:methods:entropy}, with $T_E$ and $T_F$ taken to be reasonable ``room temperature'' values of the energy/force RMSEs, respectively ($T_E = 4$ meV/atom, $T_F = 40$ meV/\AA).
The weight $\alpha$ was set to $0.2$ to resemble the higher force weighting used during training without ignoring the contributions of energy LLs in model stability.
Nevertheless, a sensitivity analysis of $S$ with respect to these parameters showed that results remained consistent around these error ranges, as long as unreasonable temperature values were not used (see Fig. \ref{fig:si:entropy_vs_temp}).

\begin{figure}[!h]
    \centering
    \includegraphics[width=0.8\linewidth]{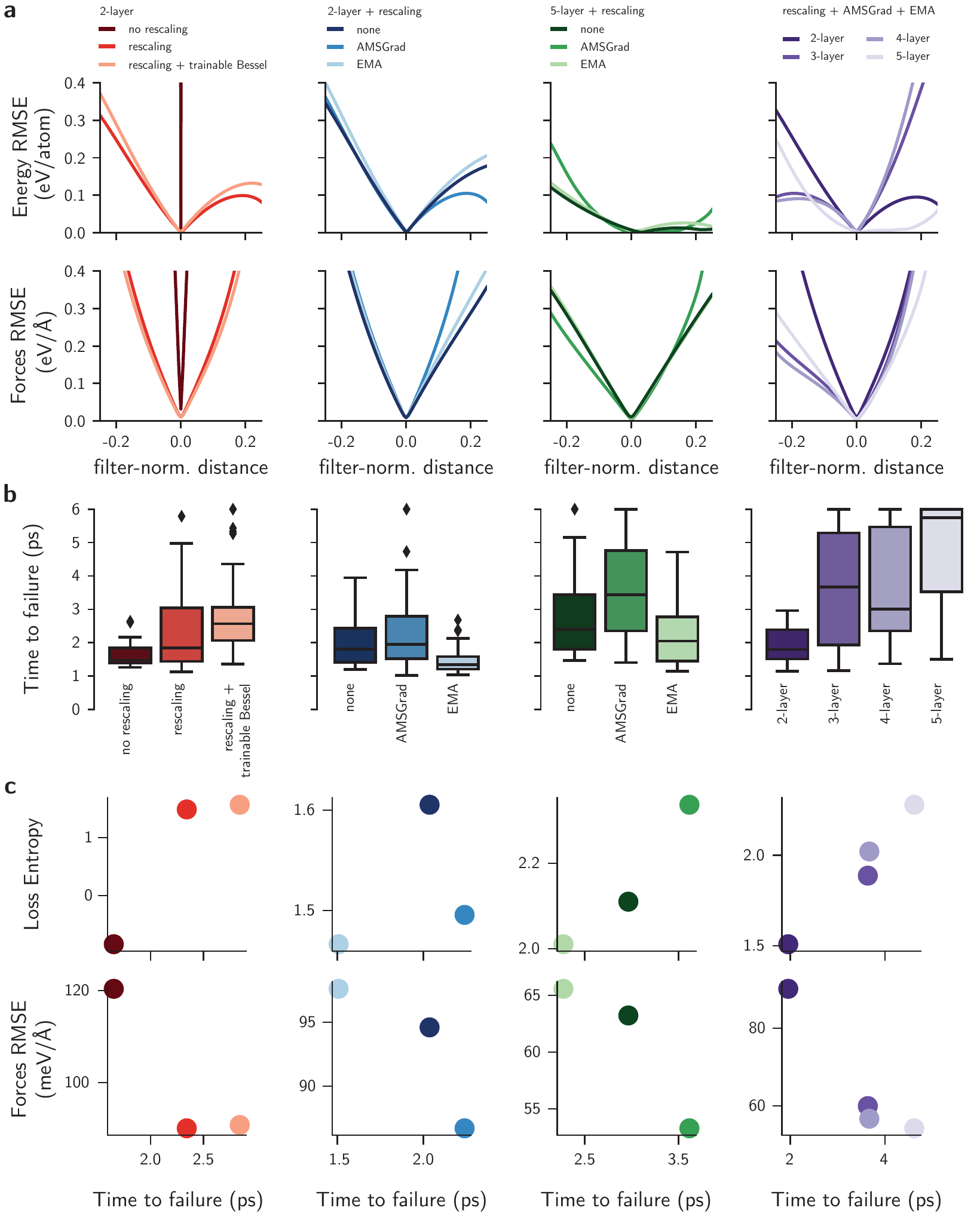}
    \caption{Analysis of LLs and MD simulation stability results for NequIP models while varying: (column 1) rescaling and Bessel weights, (column 2) optimizer settings using a model with two interaction blocks or (column 3) five interaction blocks, and (column 4) the number of interaction blocks. \textbf{a}, LLs for each model and \textbf{b}, distributions of time to failure computed over 30 MD simulations. \textbf{c}, Time to failure compared to (top) the entropy of the loss landscape (as defined in Sec. \ref{sec:methods:entropy}) or the forces RMSE on the high-temperature test sets. The horizontal line, box, and whiskers in \textbf{b} depict the median, interquartile range, and points within 150\% of the interquartile range of the distribution. Isolated diamonds are points outside of this range.}
    \label{fig:trends-nequip}
\end{figure}

The relationship between the model/training parameters, the MD stability, and LLs entropy are shown in Fig. \ref{fig:trends-nequip} (see also Table \ref{tab:si:nequip_entropy}).
The analysis is grouped into four experiments (columns in Fig. \ref{fig:trends-nequip}) to isolate the effects of specific portions of the training procedure and model architecture, thus uncovering useful trends for NNIP practitioners.
For example, considering only the distributions of time to failure from MD simulations in Fig. \ref{fig:trends-nequip}b, it can be seen that rescaling the energies predicted by the model (models ``no rescaling'' vs. ``rescaling'') greatly improves the stability of the model, especially when the Bessel weights used by the radial basis are trainable.
This behavior is reflected in the energy and forces LLs (Fig. \ref{fig:trends-nequip}a), with the LL of the model without rescaling showing considerable sharpness compared to the models using rescaled energies.
Quantitative trends are also obtained when the loss entropy and test RMSE at all splits in the 3BPA dataset are computed (Fig. \ref{fig:trends-nequip}c, first column, and Table \ref{tab:si:nequip_rmse}).
Models with rescaling showed higher average simulation time with physical trajectories, as well as increased entropy and lower forces RMSE.

When comparing how the training regime can change the extrapolation behavior of NequIP, the results show that the AMSGrad variant of the Adam optimizer \cite{Sashank2018} leads to consistent improvements in the model extrapolation both for the 2-layer and the 5-layer model (Fig. \ref{fig:trends-nequip}b, second and third column) compared to the baseline, which does not use AMSGrad. The improvements of simulation quality are particularly pronounced in the 5-layer models, where the model trained using AMSGrad shows significant improvements in simulation stability compared to the baseline despite not using trainable Bessel functions (Table \ref{tab:si:nequip_hparams}).
On the other hand, the exponential moving average (EMA) appears to degrade the extrapolation performance of the NNIPs, often leading to sharper LLs if used in isolation and with constant learning rates.
These trends are reflected in the extrapolation metrics (Fig. \ref{fig:trends-nequip}c).
Although the loss entropy underperforms compared to the forces RMSE in the case of the 2-layer models, the trend is correctly captured for the 5-layer model, where the model trained with AMSGrad showed both higher loss entropy and higher MD stability.

Finally, NequIP models can be compared according to the number of message-passing layers.
Using a fixed training regime, a higher number of layers showed pronounced improvement both in the stable simulation time (Fig. \ref{fig:trends-nequip}b), as well as in the forces RMSE and loss entropy (Fig. \ref{fig:trends-nequip}c), with the trend remaining consistent across these models.
Interestingly, the improvement in loss entropy for the 5-layer model is reflected mostly in the energy LL rather than the force LL (Fig. \ref{fig:trends-nequip}a).
Although the energy is not required when integrating the equations of motion in the NVT ensemble during an MD simulation, this improvement in LL may reflect the model's higher generalization capacity, and thus wider minima.
This phenomenon can be explained by recognizing that the out-of-domain data often results in a horizontally shifted version of the loss landscape \cite{Keskar2016}.
To explain this effect, we computed the LL of NequIP models with respect to the test set instead of the training set (see Fig. \ref{fig:si:high_t_landscapes}).
As expected, the test LL shows higher errors compared to the train LL, as seen in the vertical shift of the curves.
However, while the forces LL only undergo a vertical shift, test energy LLs also undergo a horizontal shift, indicating that the optimal model for the training set is not necessarily optimal for predicting energies of the testing set.
Furthermore, as the forces loss is often derived from the energy in NNIPs, the propagation of these mismatches may be responsible for degradation in extrapolation performance.
Thus, in general, the results in Fig. \ref{fig:trends-nequip} show that architectures and optimization strategies which result in loss landscapes with higher entropy (i.e., flatter landscapes) tend to demonstrate improved stability in MD simulations that sample out-of-domain configurations.

\subsection{Loss landscapes predict extrapolation trends in MACE}
\label{sec:results:mace}

\begin{figure}[!h]
    \centering
    \includegraphics[width=0.8\linewidth]{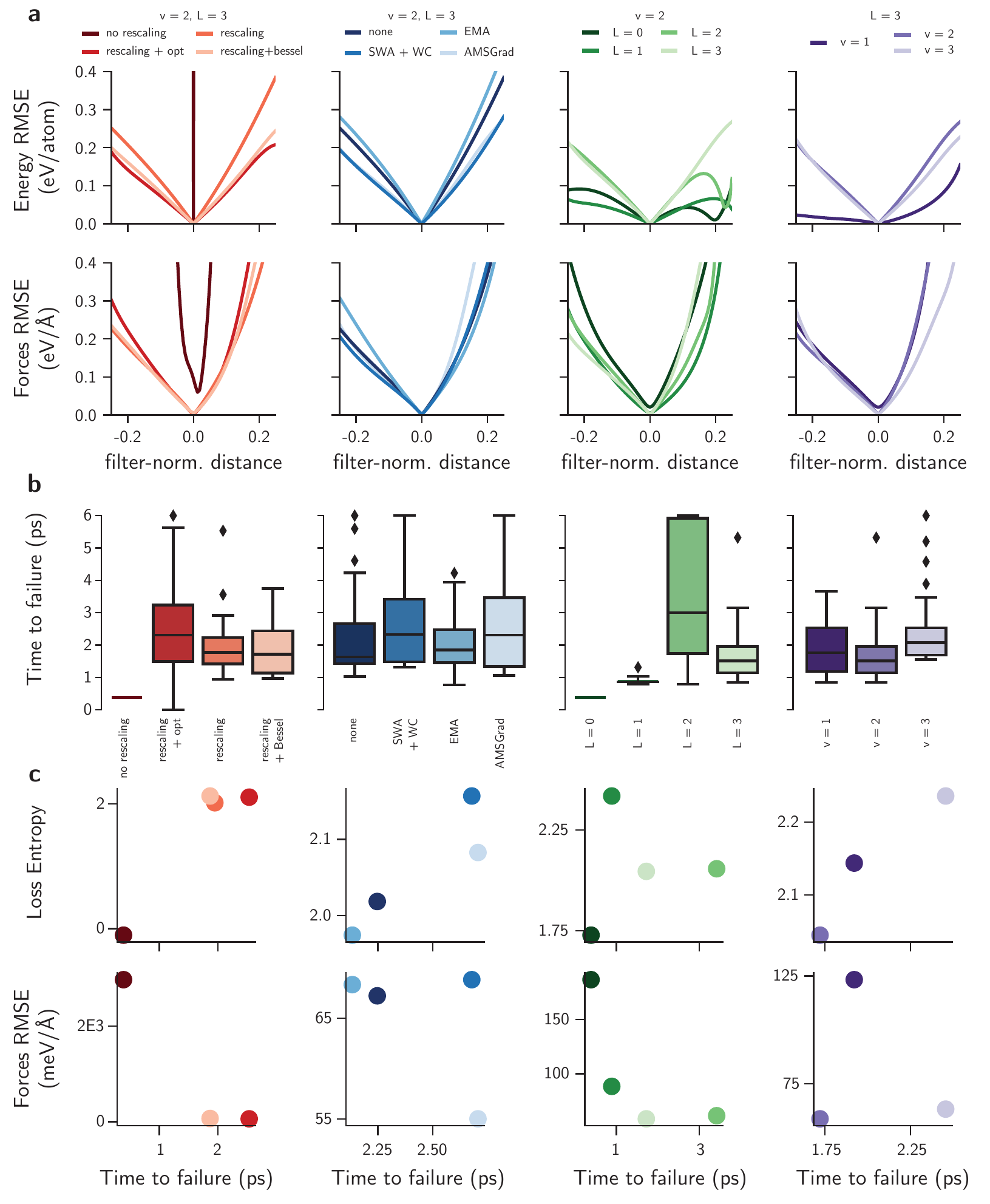}
    \caption{Analysis of loss landscapes and MD simulation stability results for MACE models while varying: (column 1) the use of rescaling; (column 2) optimizer settings using a model with a fixed body order, $v=2$, and symmetry order, $L=3$; (column 3) the symmetry order of the edge features with a fixed body order of 2; and (column 4) the body order with a fixed symmetry order of 3. \textbf{a}, Loss landscapes for each model. \textbf{b}, distributions of time-to-failure computed over 30 MD simulations. \textbf{c}, Time-to-failure compared to (top) the entropy of the loss landscape (as defined in Sec. \ref{sec:methods:entropy}) or the forces RMSE on all test sets of the 3BPA dataset. The horizontal line, box, and whiskers in \textbf{b} depict the median, interquartile range, and points within 150\% of the interquartile range of the distribution. Isolated diamonds are points outside of this range.}
    \label{fig:trends-mace}
\end{figure}

To confirm that the results from Sec. \ref{sec:results:nequip} could be extended beyond NequIP, we performed a similar study using the MACE framework.
Given the differences in architecture and available code, MACE has different hyperparameters and optimizer choices than NequIP (see Table \ref{tab:si:mace_hparams}).
Nevertheless, a similar study for MACE reproduces the trends seen for NequIP, as shown in Fig. \ref{fig:trends-mace}.

As seen in Sec. \ref{sec:results:nequip}, model stability can be greatly improved by using rescaling and AMSGrad. Also for MACE, EMA appears to lower the time-to-failure if used in isolation, with similar trends in out-of-domain RMSE, loss entropy, and MD stability seen in the NequIP case (Fig. \ref{fig:trends-mace}).
In addition, we also explored the effects of stochastic weight averaging  (SWA) \cite{Izmailov2018} and weight cycling (WC) (column 2 of Fig. \ref{fig:trends-mace}) implemented in the MACE code.
Consistent with results from the use of SWA for image classification tasks \cite{Izmailov2018}, SWA + WC is shown to improve model generalizability, leading to higher simulation times and flatter loss landscapes.
Interestingly, the model trained with SWA + WC exhibits similar force RMSE in high-temperature samples compared to the baseline, but a higher loss entropy and higher MD stability (Fig. \ref{fig:trends-mace}c).
As SWA flattens the loss landscape by design \cite{Izmailov2018}, implementing this strategy in other NNIP models may be a low-cost modification for improving the generalization capacity of different architectures.
The use of WC can also help the optimization process (Sec. \ref{sec:results:ll}) and lead to better energy landscapes overall.

For the MACE study, we also analyzed the relationship between the bond order of the model (column 4) and their stability in production MD simulations.
In this case, the loss entropy recovers the trend in MD stability better than the forces RMSE (Fig. \ref{fig:trends-mace}c, column 4).
Interestingly, despite the higher errors of the $v=1, L=3$ model compared to its $v=2, L=3$ counterpart, the former still exhibits a higher average time to failure.
This observation can be traced back to the flatter energy LL (Fig. \ref{fig:trends-mace}a, column 4), as also seen in the case of NequIP.

Finally, the models were compared according to the symmetry order of the edge features (column 3).
In general, increasing the value of $L$ from $L=0$ to $L=2$ led to more stable simulations, although the model with $L=3$ shows a degradation of the production behavior.
Nevertheless, the LLs do not follow the expected trends explained so far in this work.
We believe the filter normalization technique discussed in Sec. \ref{sec:methods:ll} cannot provide comparable LLs upon changes of $L$ in MACE.
As this parameter affects the tensor order of some model components \cite{Batatia2022}, the number of parameters in these particular filters grows with $\mathcal{O}(N^L)$, which may be affecting the filter normalization.
In contrast, adding message-passing layers with similar numbers of parameters, as in NequIP, allows the filter normalization technique to create comparable loss landscapes.
This result shows some limitations of the loss entropy when comparing models with different architectures or hyperparameters, and may require different strategies for computing them in the future.

\subsection{Loss landscapes and data efficiency}
\label{sec:results:all}

As extrapolation power and data efficiency are conceptually related, a natural extension of Secs. \ref{sec:results:nequip} and \ref{sec:results:mace} is to verify whether the loss landscape entropy can also be used to predict the data efficiency of a model during training.
As with many other applications of deep learning, generating training data is often one of the most costly steps in the development process.
Especially considering the ``denoising'' effects of NNIPs demonstrated in Sec. \ref{sec:results:representation}, identifying training techniques and model architectures that lead to more data-efficient training has the potential to greatly reduce the computational cost of building new NNIPs.

\begin{figure}[!h]
    \centering
    \includegraphics[width=\linewidth]{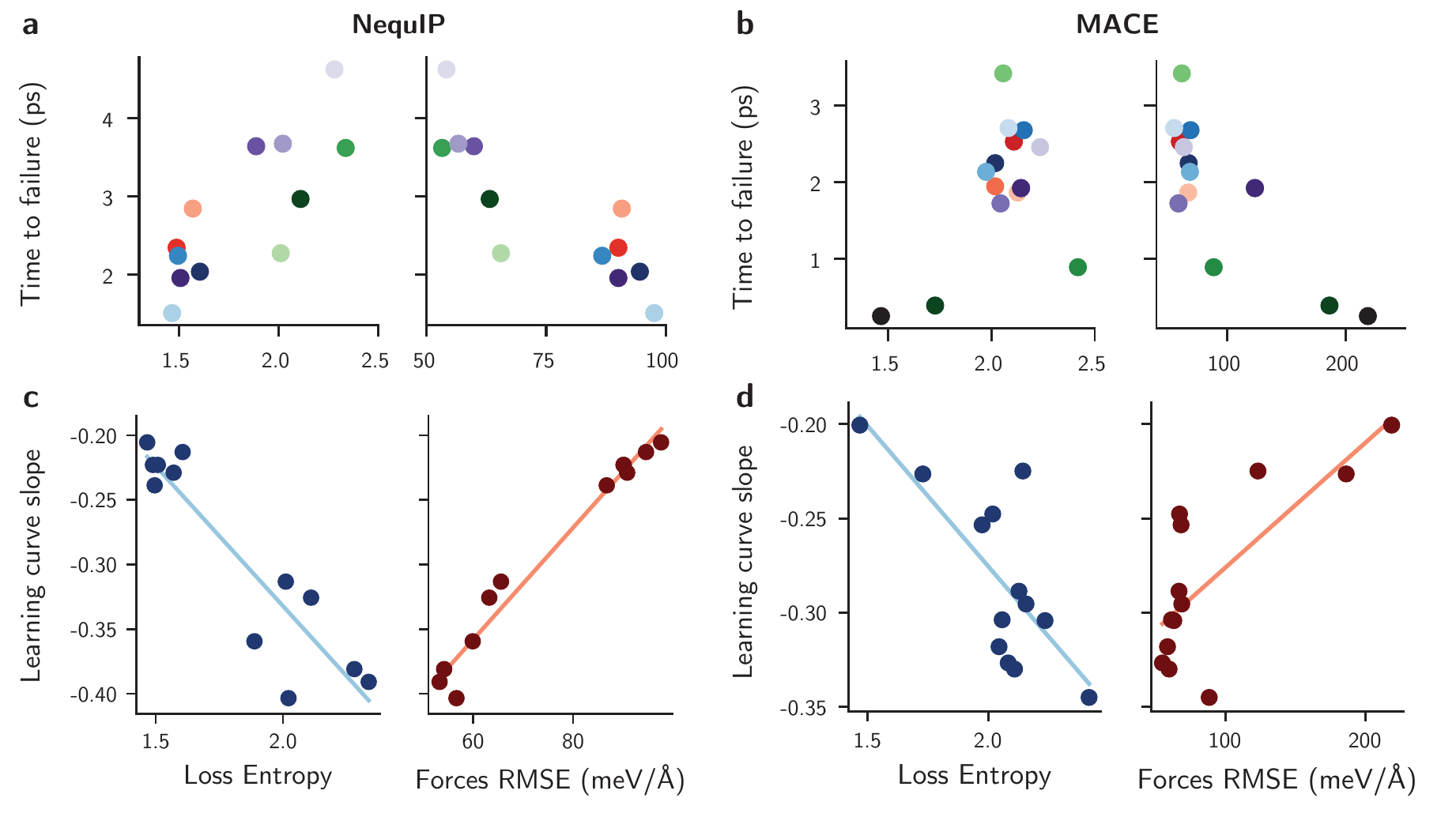}
    \caption{Loss landscape entropy and high-temperature force test RMSEs compared to \textbf{a,b} MD time-to-failure or \textbf{c,d} learning curve slope for \textbf{a,c} NequIP and \textbf{b,d} MACE models. The points . Learning curve slopes were computed as described in Sec. \ref{sec:results:all}. Colors for points in \textbf{a} and \textbf{b} were chosen to match the colors in Fig. \ref{fig:trends-nequip} and Fig. \ref{fig:trends-mace} respectively. Lines in \textbf{c} and \textbf{d} represent linear fits to the data.}
    \label{fig:trends-all}
\end{figure}

To test whether loss entropy can predict the data efficiency of models, we computed the learning curves for NequIP and MACE models with the hyperparameters selected in Sections \ref{sec:results:nequip} and \ref{sec:results:mace}.
This is achieved by training models with the specified training parameters and architectures to the same subsets of the 3BPA training set using only 25, 125, 250, or 500 samples from the 300 K split (Tables \ref{tab:si:nequip_rmse} and \ref{tab:si:mace_rmse}).
Following previous work \cite{Batzner2022}, the slopes of the learning curves were then computed by fitting the line $\log{n} = m \log{\varepsilon} + b$ to the number of training samples $n$ and the force RMSEs $\varepsilon$ calculated using all test splits (300, 600, and 1200 K), and comparing the slopes $m$.
Consistent with the results from Sections \ref{sec:results:nequip} and \ref{sec:results:mace} (combined in Figures \ref{fig:trends-all}a,b), the high-temperature force RMSEs (i.e., the points from the learning curves using $n=500$) are good predictors of the learning curve slope.
In addition, Figs. \ref{fig:trends-all}c,d show that the loss entropy also has a high correlation with the slope of the learning curve despite being computed using only the training set.
As discussed in Sec. \ref{sec:results:mace}, these correlations are less explicit for the case of MACE models, and thus show that none of the metrics is truly universal for predicting stability in simulations and learning curve slopes.
Nevertheless, these results further demonstrate that loss landscapes provide information on the extrapolation behavior of NNIPs despite being derived only from the training set.

\section{Conclusion}

In this work, we motivate the need for additional metrics beyond RMSE by showing that in-domain errors fail to predict model stability in the high-accuracy regime despite large-scale trends in extrapolation behavior and NNIP robustness to noise.
We propose the use of loss entropy as a metric for quantifying the extrapolation behavior, and demonstrate that it correlates well with out-of-distribution error and stability in production simulations.
Using large studies with NequIP and MACE, we show how models containing flatter loss landscapes exhibit better extrapolation behavior, and how different training parameters can be used to achieve them.
For example, rescaling, AMSGrad, and SWA were shown to increase the loss entropy and MD stability, and may be important tools when training NNIP models.
Similarly, models with similar test error can be distinguished by their energy loss landscape, with models displaying broader minima performing better in extrapolation tasks.
Future studies can address shortcomings of loss landscape visualizations in NNIP by analyzing how filter-normalization can be made more suitable for NNIP architectures, or to better isolate the effects of key architectural changes in the loss.
Nevertheless, these results can better inform the development of new model architecture and optimization strategies in NNIPs, facilitating their use in general materials simulation.

\section*{Code Availability}

The package \texttt{ip\_explorer} and the additional code/data used to reproduce the results of this paper will be made available after internal review at LLNL. This preprint will be updated with the relevant links. Loss landscape calculations were performed using the the code from the public package \url{https://github.com/marcellodebernardi/loss-landscapes}. Training codes for NequIP and MACE are available from their original authors as described in Appendix \ref{sec:si:training}.

\section*{Data Availability}
The datasets used to train the models in this work were obtained directly from their original sources. For convenience, we provide the links to each source here: \url{https://github.com/davkovacs/BOTNet-datasets/tree/main/dataset_3BPA} (3BPA) and \url{https://github.com/atomistic-ml/ani-al} (ANI-Al).

\section*{Author Contributions}

\textbf{Joshua Vita:} Methodology, Software, Validation, Investigation, Data Curation, Writing - Original Draft, Writing - Review \& Editing, Visualization. 
\textbf{Daniel Schwalbe-Koda:} Conceptualization, Methodology, Software, Validation, Investigation, Data Curation, Writing - Original Draft, Writing - Review \& Editing, Visualization, Supervision.

\section*{Acknowledgments}
This work was performed under the auspices of the U.S. Department of Energy by Lawrence Livermore National Laboratory under Contract DE-AC52-07NA27344, funded by the Laboratory Directed Research and Development (LDRD) Program at LLNL under project tracking code 22-ERD-055. The authors thank Vincenzo Lordi, the Quantum Simulations Group at LLNL, and Simon Batzner for the discussions. We also thank Ilyes Batatia for the support with the MACE code.

Manuscript released as \texttt{LLNL-JRNL-845001-DRAFT}.


\appendix

\renewcommand{\thefigure}{S\arabic{figure}}
\setcounter{figure}{0}
\renewcommand{\thetable}{S\arabic{table}}
\setcounter{table}{0}

\section{Training details}
\label{sec:si:training}

All models were trained on a single node of the Lassen supercomputer, using one NVIDIA V100 (Volta) GPU. Most models were trained for a maximum walltime of 12 hours, though some models were trained for an additional 12 hours to ensure convergence. For all models, an energy:force weight of 1:1000 was used on the 3BPA dataset, and 1:10 on ANI-Al. A cutoff distance of 5.0 \AA~ was used for all models, though the effective interaction distance may have varied depending on the number of interaction blocks used, as described below.

\subsection{NequIP}

The NequIP model was trained using its NequIP package \cite{Batzner2022}, version 0.5.5 (\url{https://github.com/mir-group/nequip}), implemented in PyTorch. NequIP model architecture and training parameters were chosen to match those used in \cite{Batatia2022}, with some modifications as specified in this paper. See Table \ref{tab:si:nequip_hparams} for an overview of key architecture and optimizer modifications used in this work. The Adam optimizer was used with an initial learning rate of 0.005, and the \texttt{ReduceLROnPlateau} scheduler with a patience of 50 and a decay factor of 0.5. Any other parameters used the defaults specified by \url{https://github.com/mir-group/nequip/blob/main/configs/example.yaml}, version 0.5.5.

\subsection{MACE}

The MACE code was trained using its public package (\url{https://github.com/ACEsuit/mace}), version 0.1.0. The MACE code required a minor patch to allow for the Bessel weights to be made trainable in one of the studies. MACE model training parameters were chosen to match those used in \cite{Batatia2022}, with modifications as specified in this paper. See Table \ref{tab:si:mace_hparams} for an overview of key architecture and optimizer modifications used in this work. The Adam optimizer was used with an initial learning rate of 0.005, and the \texttt{ReduceLROnPlateau} scheduler with a patience of 50 and a decay factor of 0.8. Any other parameters used the defaults specified by \url{https://github.com/ACEsuit/mace}, version 0.1.0.

\section{MD simulations}
\label{sec:si:md}

\begin{figure*}[!htb]
    \centering
    \includegraphics[width=0.7\linewidth]{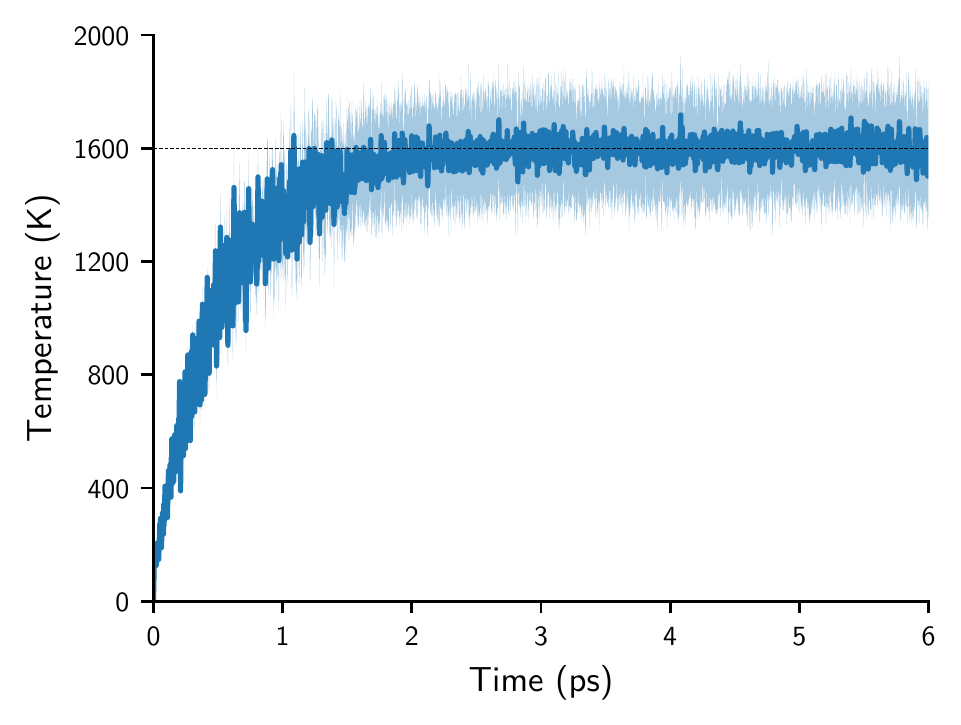}
    \caption{Average temperature of the 36 MD simulations performed with NequIP that did not demonstrate unphysical behavior throughout the 6 ps of total simulation time. The blue line is the average temperature for each of the time step, and the shaded area is the standard deviation of temperatures for each time step during these 36 different trajectories.}
    \label{fig:si:md_temperature}
\end{figure*}

Molecular dynamics simulations were performed for the 3BPA molecule using each of the models under study. In total, 30 trajectories were performed per model to obtain better statistics of their production behavior. Simulations were performed in the NVT ensemble using the Berendsen thermostat \cite{berendsen1984molecular} implemented in ASE \cite{ase2017}. The initial configuration used for the simulation was chosen to be the ground state of the 3BPA molecule. The simulation was performed at 1600 K and a timestep of 1 fs to force the models to evaluate configurations outside of their training set (300 K samples) and beyond those quantified by the test set errors (1200 K samples). The time constant of the Berendsen thermostat was chosen to be 250 fs. The simulation was performed for 6 ps for all models. This time length was sufficient to ensure the temperature was equilibrated and remained at 1600 K for about 4 ps (see Fig. \ref{fig:si:md_temperature}).

MD trajectories were considered unphysical if the distance between bonded atoms increased above 2 \AA. This bond rupture was the main source of failure observed in the trajectories, and thus is the only one considered in this work.

\clearpage
\section{Toy example on linear regressor trained on noisy data}
\label{sec:linregress}

To illustrate how redundancy in data points can help recover the underlying data despite the noise, we provide an example with a linear regressor trained on noisy data.
A linear function can be obtained by fitting two parameters to the dataset, thus requiring at least two data points to be fit.
When more data points are available, a least-squares method can be employed to obtain the only solution that minimizes the error of the fit towards the dataset.
Because linear models are not robust to outliers, this pedagogical example helps illustrate how redundant  data helps recover the underlying generative function despite the noise.
To illustrate this effect, we consider a linear function $y$ given by

\begin{equation}\label{eq:si:linregress}
    y = 2x + 1,
\end{equation}

\noindent which represents the ground truth data. Using this definition, the ``corrupted'' data $\Tilde{y}$ is given by

\begin{equation}\label{eq:si:noisy_y}
\Tilde{y} = y + \sigma \varepsilon
\end{equation}

\noindent where $\sigma$ is a parameter and $\varepsilon$ is sampled from a normal distribution with mean zero and variance one.

A noisy dataset $\{(x_i, \Tilde{y}_i)\}$ containing N elements is constructed by taking N linearly spaced values of $x$ in the interval $[0, 1]$ and computing the value of $\Tilde{y}$ using Eq. \eqref{eq:si:noisy_y}. Then, this noisy dataset is used to fit a linear regression model $\hat{y}$ expressed as

\begin{equation}\label{eq:si:linregress_model}
    \hat{y} = Ax + B,
\end{equation}

\noindent where the linear model is trained with the least squares method without regularization. The model error against the true data is then computed as the RMSE between the predictions $\hat{y}$ and the true data $y$.

\begin{figure*}[!htb]
    \centering
    \includegraphics[width=\smallfigwidth]{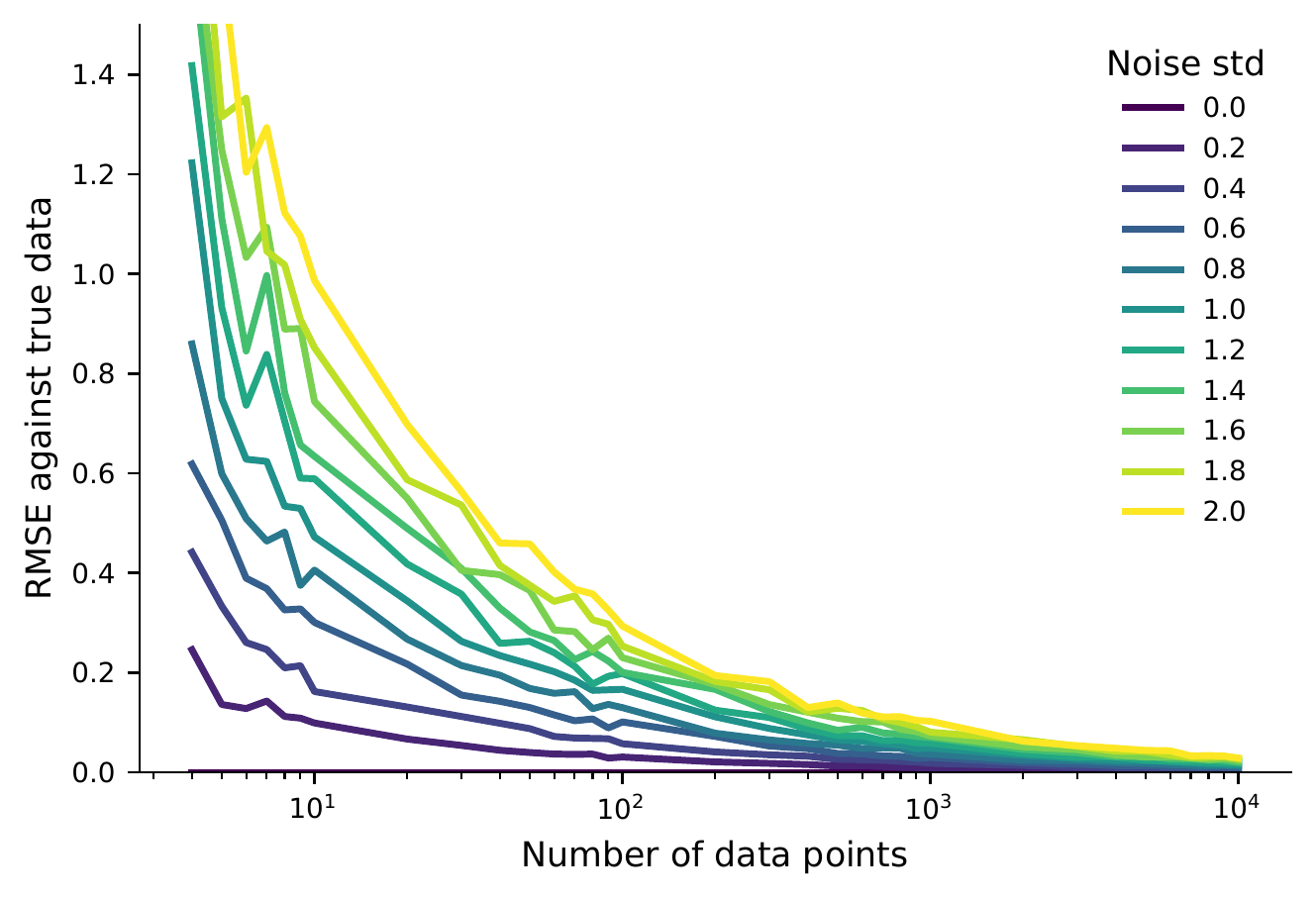}
    \caption{Average RMSE of linear regressor models trained on a noisy linear dataset. The RMSE of each model is computed against the true linear function, even though the model was trained on a noisy dataset. The average RMSE is obtained by performing this experiment 100 times for each dataset size and noise level. The noise std corresponds to $\sigma$ in Eq. \eqref{eq:si:noisy_y}.}
    \label{fig:si:linregress_noise}
\end{figure*}

Figure \ref{fig:si:linregress_noise} shows the results of fitting the linear model to the noisy data. At zero noise, the RMSE is always zero, as the linear fit recovers the true function $y$. However, as noise is added to the system, the average RMSE against the true dataset increases, particularly in the low-data regime. As more data is added to the system, the effects of the noise are compensated by the redundancy of the data points in conveying the true function of the system. For a dataset with more than 1000 data points in this linear regression example, even high levels of noise (in the units of Eq. \eqref{eq:si:linregress}) are compensated by enough redundancy in data.

\clearpage

\section{Additional figures}

\subsection{3BPA}
\label{sec:si:additional_3bpa}

The figures in this section were all generated using the 3BPA dataset, either with the raw data directly or with a model trained to it.

\begin{figure}[!htb]
    \centering
    \includegraphics[width=0.8\linewidth]{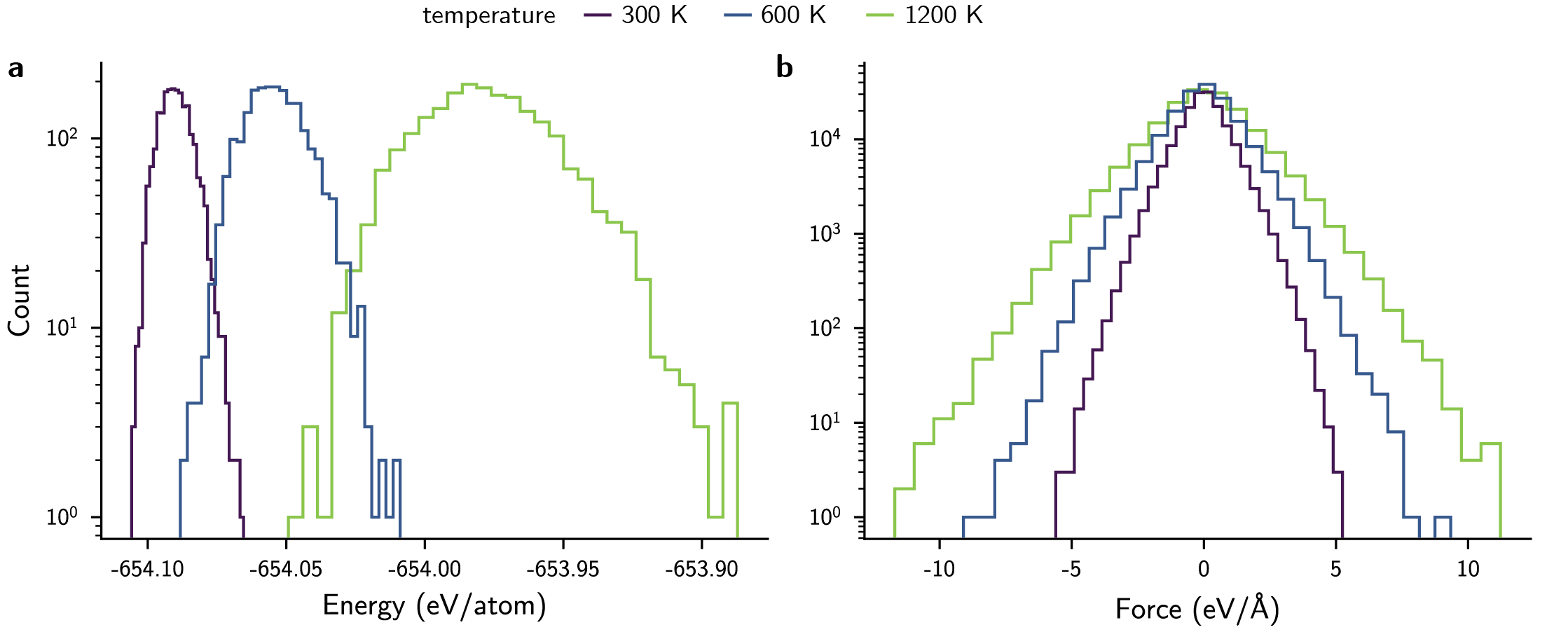}
    \caption{Distributions of \textbf{a}, energies and \textbf{b}, forces for the 3BPA dataset at the three sampling temperatures. To corrupt the labels, the standard deviation of these distributions are computed, leading to $\sigma_\mathrm{DFT} = 6.12$ meV/atom for energies and $\sigma_\mathrm{DFT} = 0.95$ eV/\AA~ for forces. Noise in forces are added to each component of the force, for each atom in the system.}
    \label{fig:si:hist_3bpa}
\end{figure}

\begin{figure*}[!htb]
    \centering
    \includegraphics[width=\smallfigwidth]{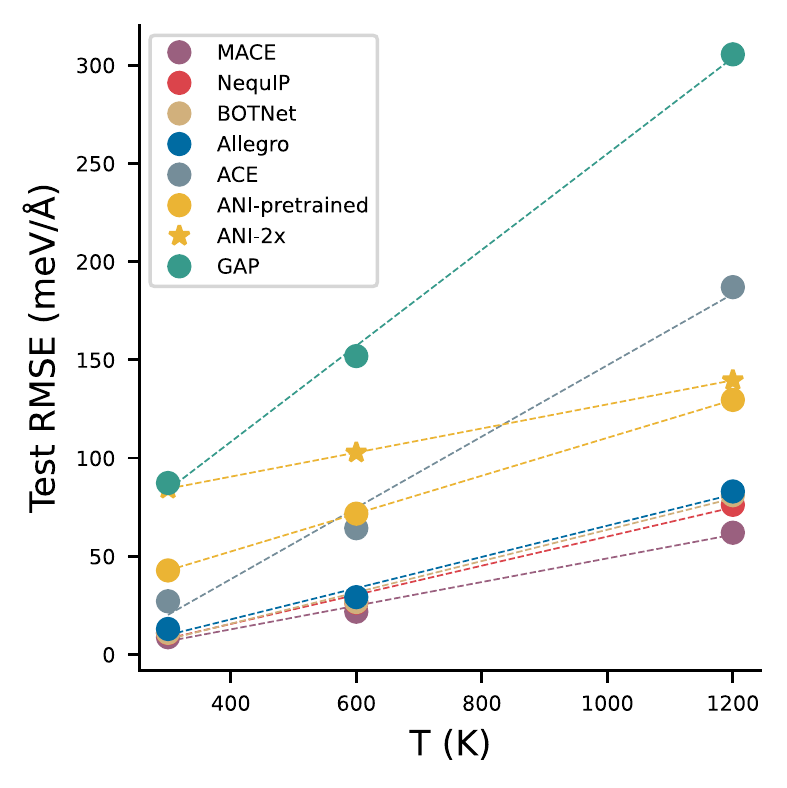}
    \caption{Linear fits to the 3BPA testing set error, as described in Sec. \ref{sec:results:representation}}
    \label{fig:si:generalizability_fits}
\end{figure*}

\begin{figure*}[!htb]
    \centering
    \includegraphics[width=0.7\linewidth]{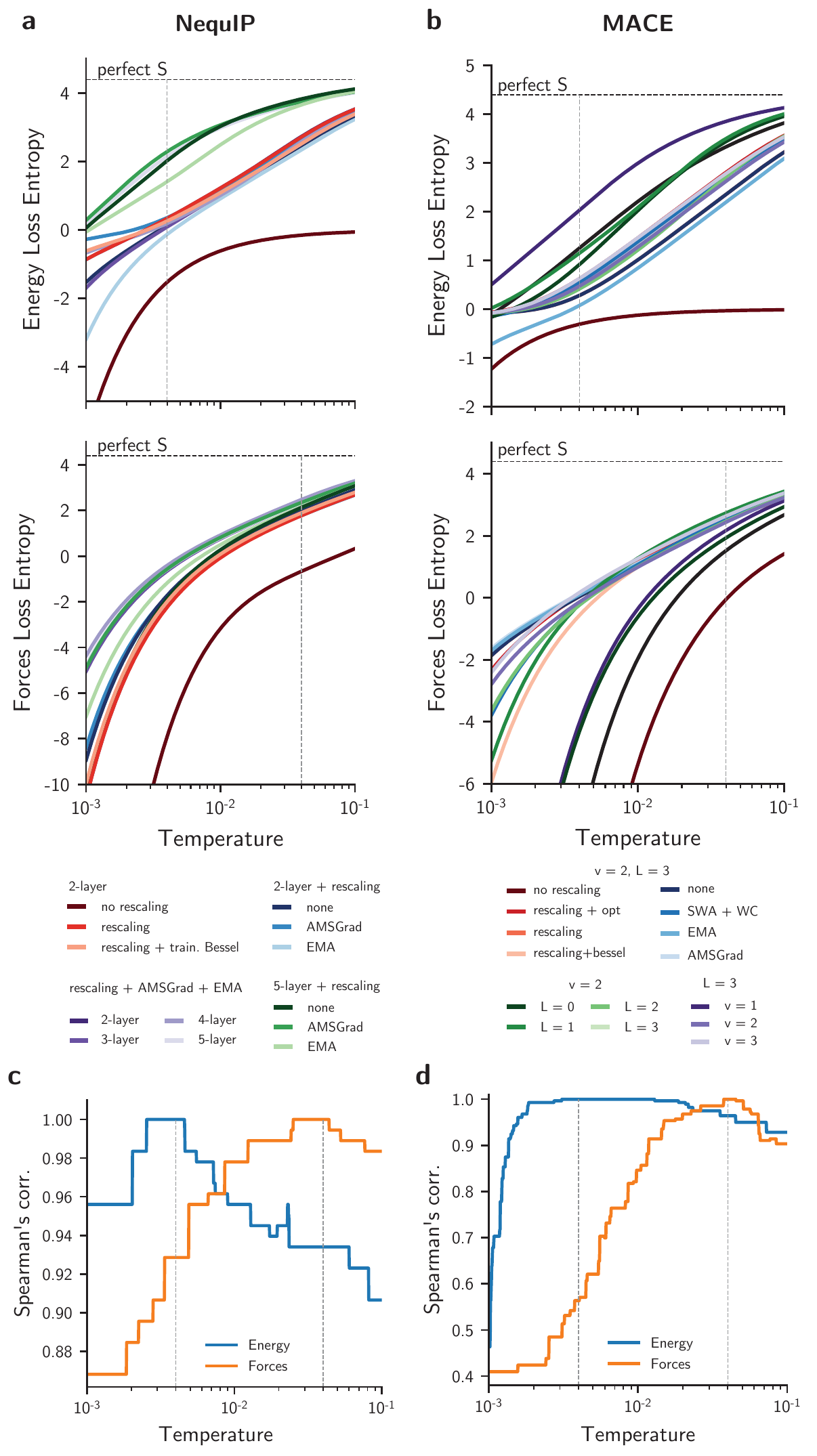}
    \caption{Sensitivity analysis of LL entropy with respect to the ``temperature'' for all \textbf{a}, NequIP and \textbf{b}, MACE models in this study. The horizontal dashed line indicates the entropy of a perfectly flat loss of zero. Although the magnitude of the entropy changes substantially, the ranking of the models according to their entropy remains largely constant around the values of temperature $T_E = 4$ meV/atom and $T_F = 40$ meV/\AA~ chosen in this study (vertical dashed lines), as shown in \textbf{c}, \textbf{d} for NequIP and MACE, respectively.}
    \label{fig:si:entropy_vs_temp}
\end{figure*}

\begin{figure*}[!htb]
    \centering
    \includegraphics[width=\linewidth]{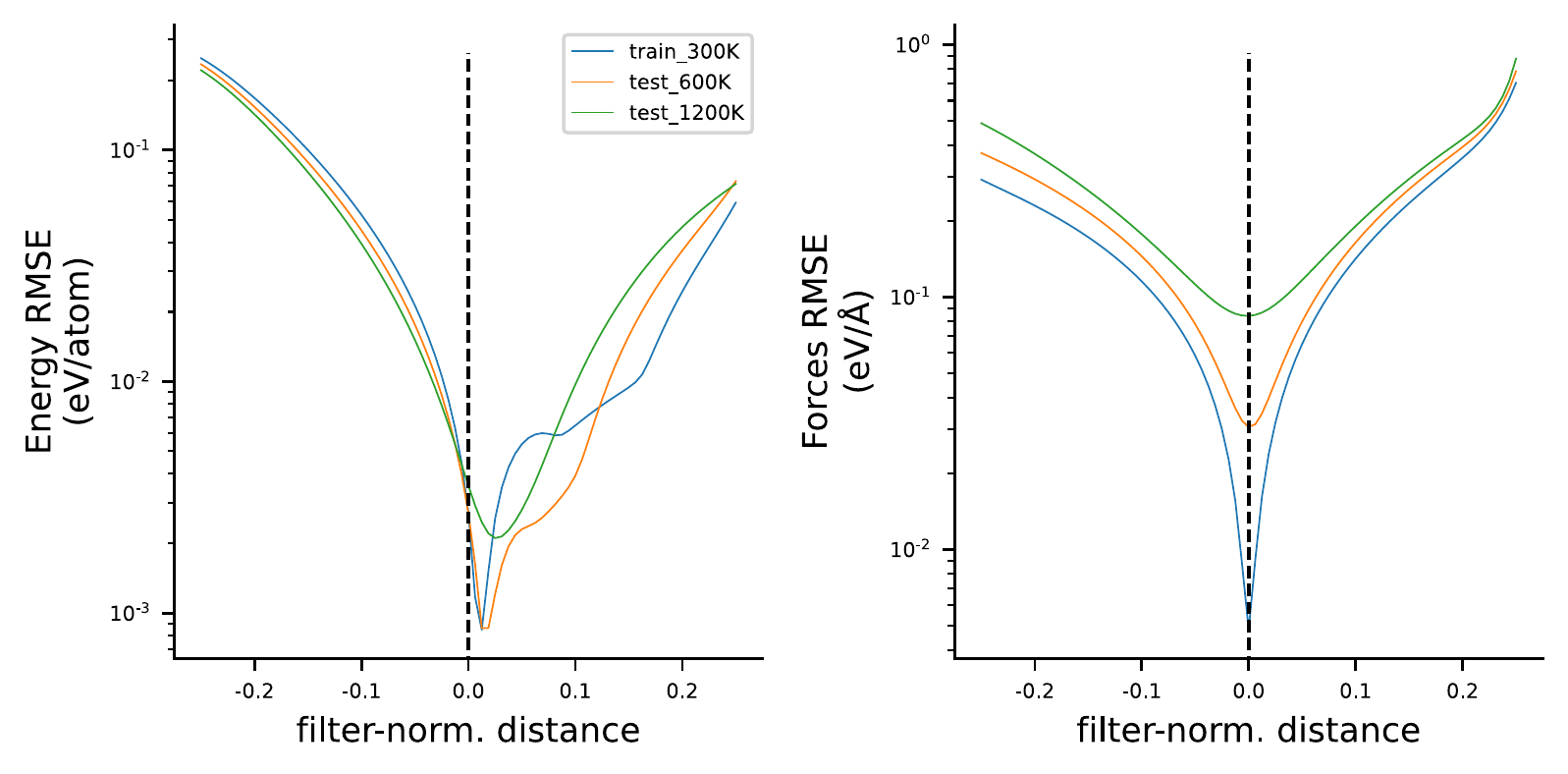}
    \caption{1D loss landscapes for the 5-layer ``rescaling + AMSGrad + EMA'' NequIP model from Sec. \ref{sec:results:nequip} using the original 300 K training set, or the 600 K and 1200 K test sets. While the energy landscapes demonstrate a horizontal shift with increasing temperature, as theorized by \cite{Keskar2016}, no such shift is observed in the force landscapes.}
    \label{fig:si:high_t_landscapes}
\end{figure*}

\begin{figure}[!htb]
    \centering
    \includegraphics[width=0.7\linewidth]{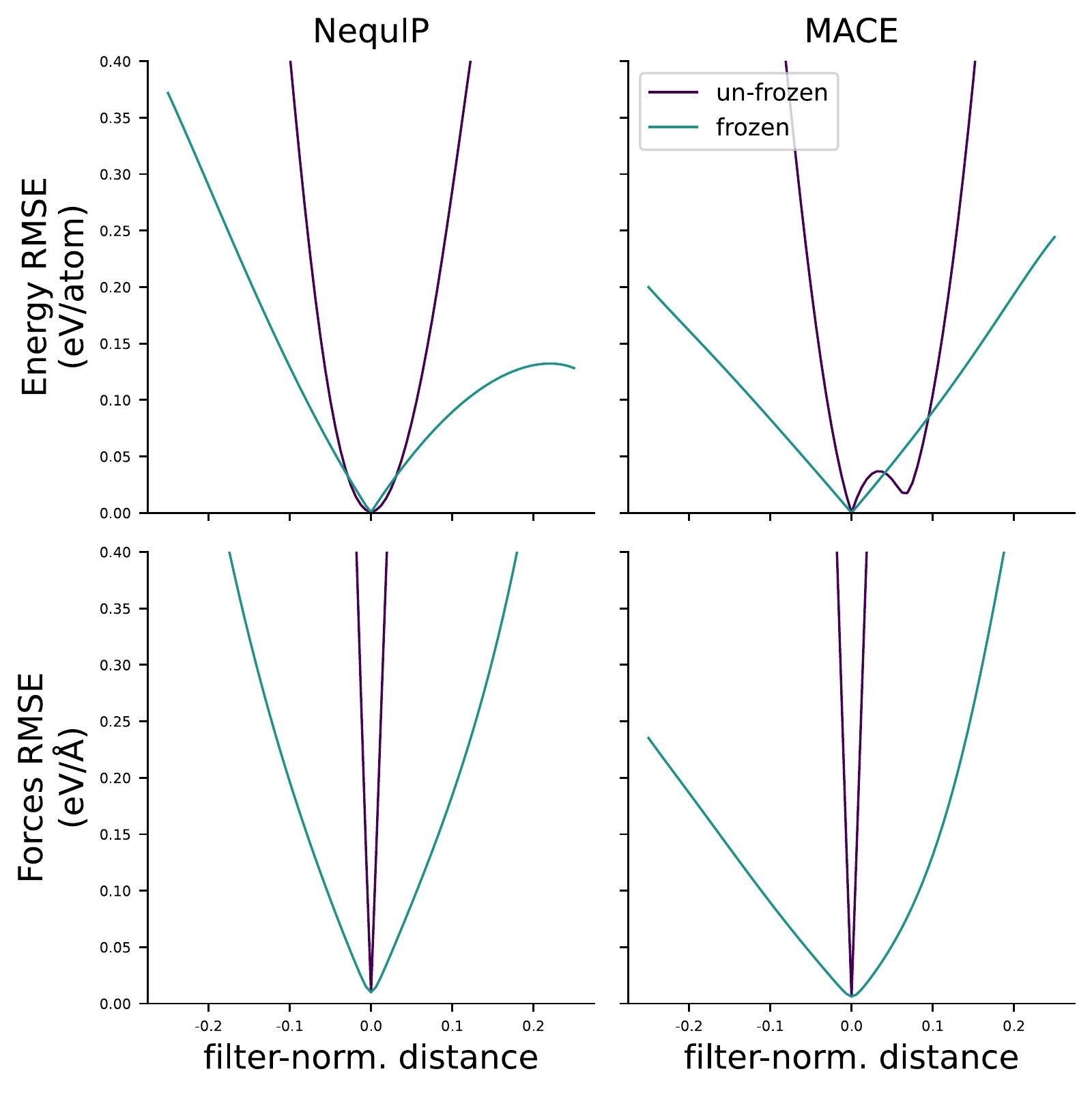}
        \caption{``Frozen'' 1D loss landscapes for NequIP and MACE models with trainable Bessel weights. The ``un-frozen'' landscapes were generated as described in Sec. \ref{sec:methods:ll}. The ``frozen'' landscapes were generated by fixing the Bessel weights to their optimized value (corresponding to the model at a filter-normalized distance of 0) when generating the loss landscapes. With the exception of this figure, all LLs shown in this work correspond to ``frozen'' LLs.}
    \label{fig:si:ray_tracing_frozen}
\end{figure}

\begin{figure*}[!htb]
    \centering
    \includegraphics[width=\smallfigwidth]{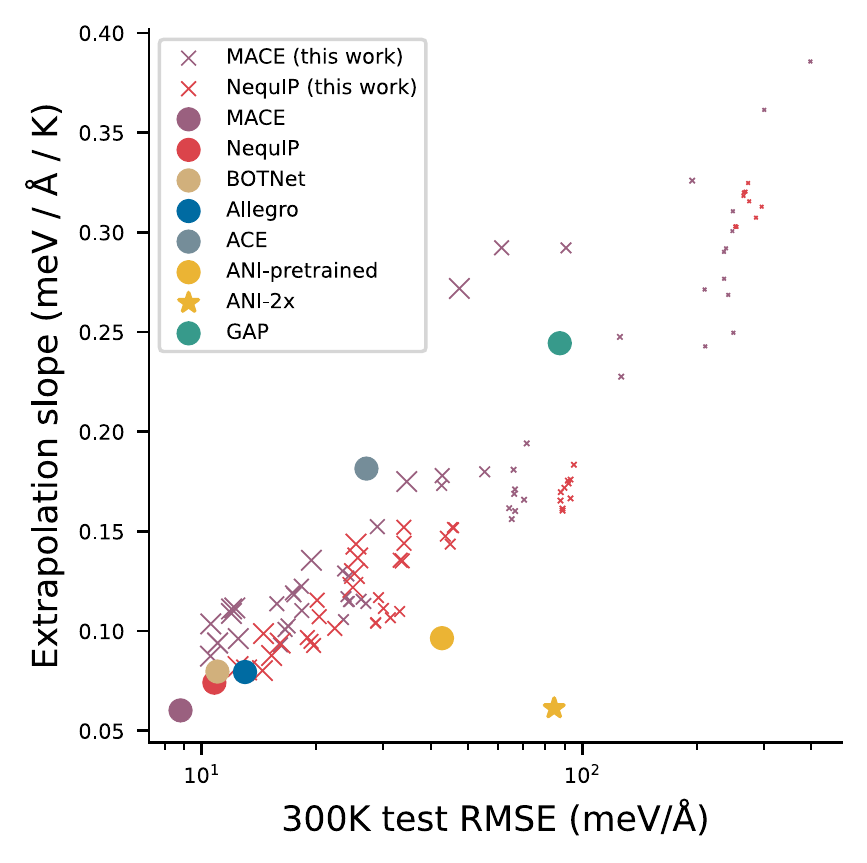}
    \caption{Plot shown in Fig. \ref{fig:noisy_fits}c, but including all models from this work. As many of the models used to generate the learning curves from the main text were trained to fewer than 500 configurations (the size of the full 3BPA training dataset), we scale the corresponding points in this plot by the number of 3BPA configurations used for training (ranging from 5 to 500). Despite the vertical dispersion, models still exhibit a rough scaling relation proposed in Fig. \ref{fig:noisy_fits}c, with smaller training sets exacerbating the extrapolation slope in the high-error regime.} 
    \label{fig:si:error_vs_slope_literature_with_ablation}
\end{figure*}

\clearpage

\subsection{ANI-Al}
\label{sec:si:additional_ani}

In addition to the 3BPA dataset, part of this work also uses the aluminum dataset from Smith et al. \cite{Smith2021} (ANI-Al).
The ANI-Al dataset was generated using active learning intended to thoroughly sample configurational space, and was originally used to construct a model for running shock simulations.
This additional dataset was chosen for being a condensed matter system with wide configurational diversity but restricted composition, thus being a strong contrast to the molecular 3BPA.

The figures in this section were all generated using the ANI-Al dataset, either with the raw data directly or with a model trained to it.

\begin{figure}[!htb]
    \centering
    \includegraphics[width=0.8\linewidth]{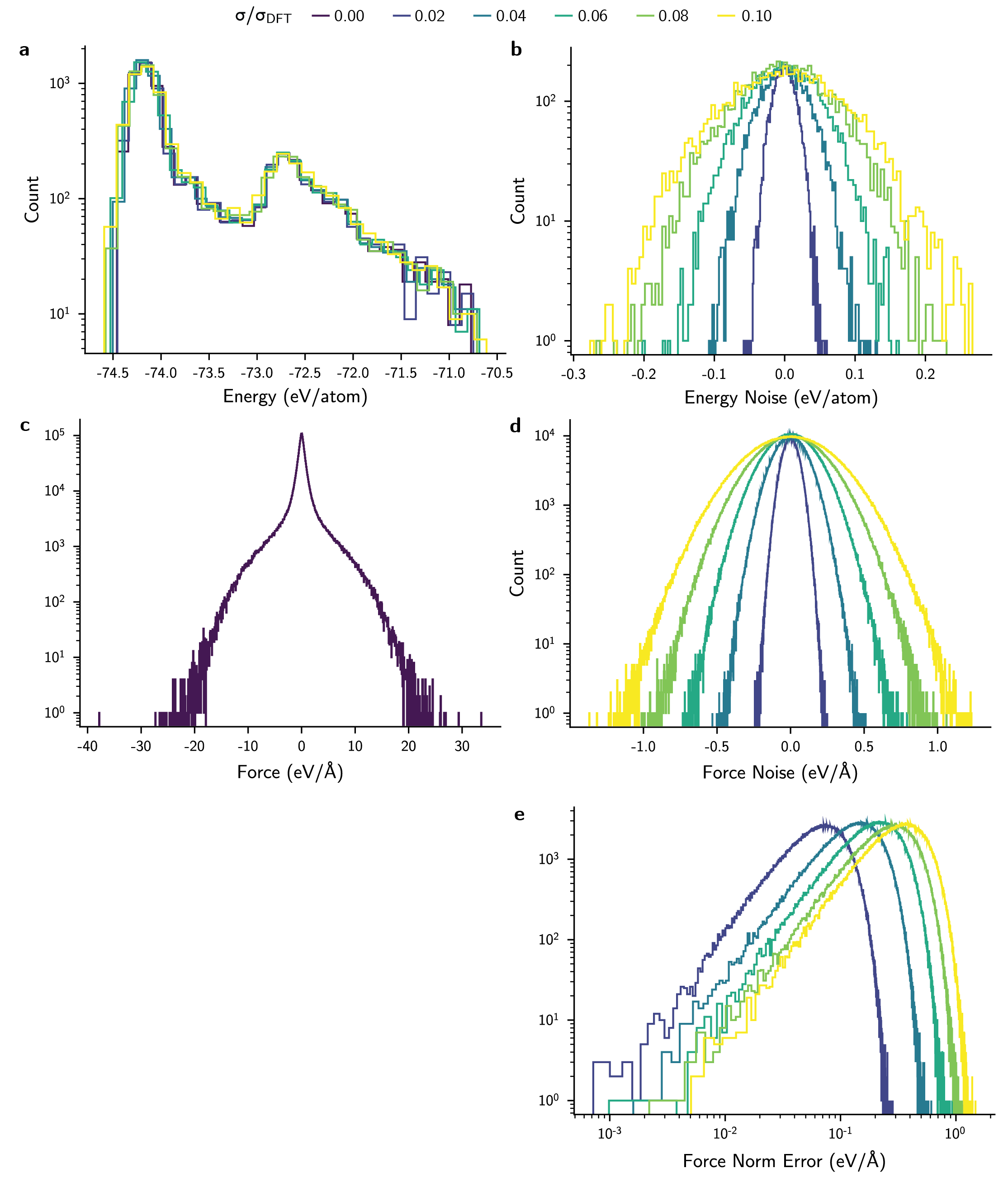}
    \caption{Distributions of \textbf{a}, energies, \textbf{b}, energy noises, \textbf{c}, forces, and \textbf{d-e}, force errors for the ANI-Al dataset. For \textbf{d-e}, the distributions of deviations are taken with respect to each force component for the noisy datasets. The deviation is computed for each coordinate of each force vector. \textbf{e}, shows the distributions of errors of the final norm of the noisy forces compared to the original ones.}
    \label{fig:si:hist_ani}
\end{figure}

\begin{figure}[!htb]
    \centering
    \includegraphics[width=0.8\linewidth]{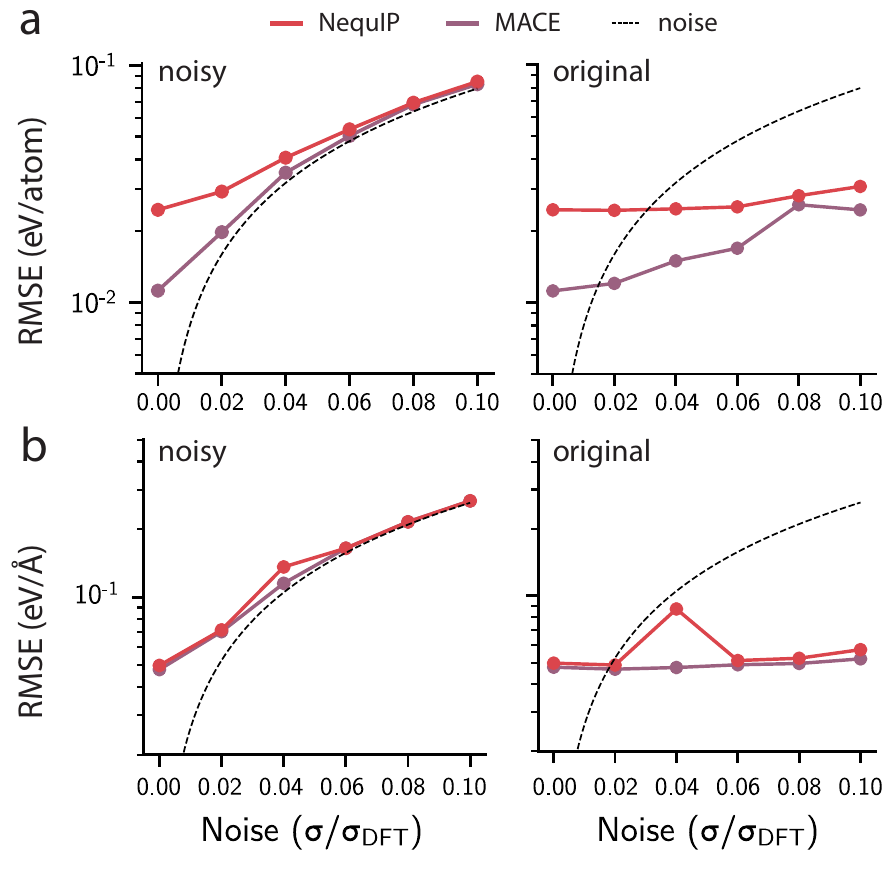}
    \caption{RMSE of \textbf{a} energies (eV/atom) and \textbf{b} forces (eV/\AA) for models trained to noisy versions of the ANI-Al dataset.
    The models trained on noisy data are then evaluated on their noisy training set (``noisy'' chart) or on the original, uncorrupted dataset (``original''). The black dashed lines correspond to the amount of noise that was added to the DFT energies in units of eV/atom. Distributions of \textbf{b}, energies and \textbf{c}, forces noises for this high-noise ANI-Al dataset.}
    \label{fig:si:noisy_fits_ani}
\end{figure}

\begin{figure}[!htb]
    \centering
    \includegraphics[width=\smallfigwidth]{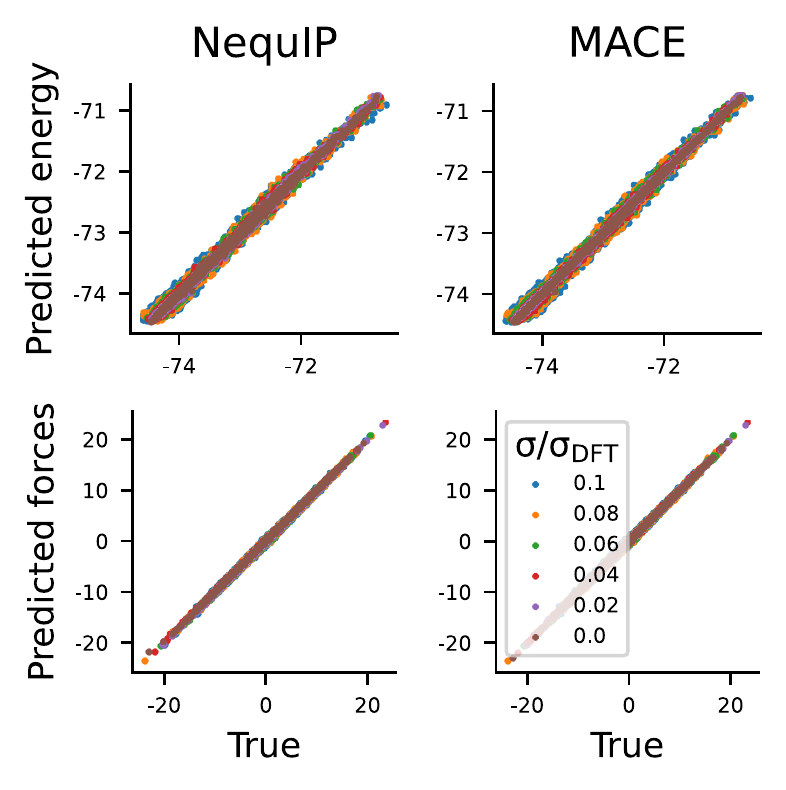}
    \caption{Parity plots for all four models trained to the ANI-Al dataset with increasing noise levels. The points for each noise level are plotted on top of each other, with colors based on the amount of noise as specified in the legend, emphasizing that the models are learning predictions that are very similar to the ones that they make when trained to the non-noisy dataset.}
    \label{fig:si:true_pred_scatter}
\end{figure}

\begin{figure}[!htb]
    \centering
    \includegraphics[width=0.8\linewidth]{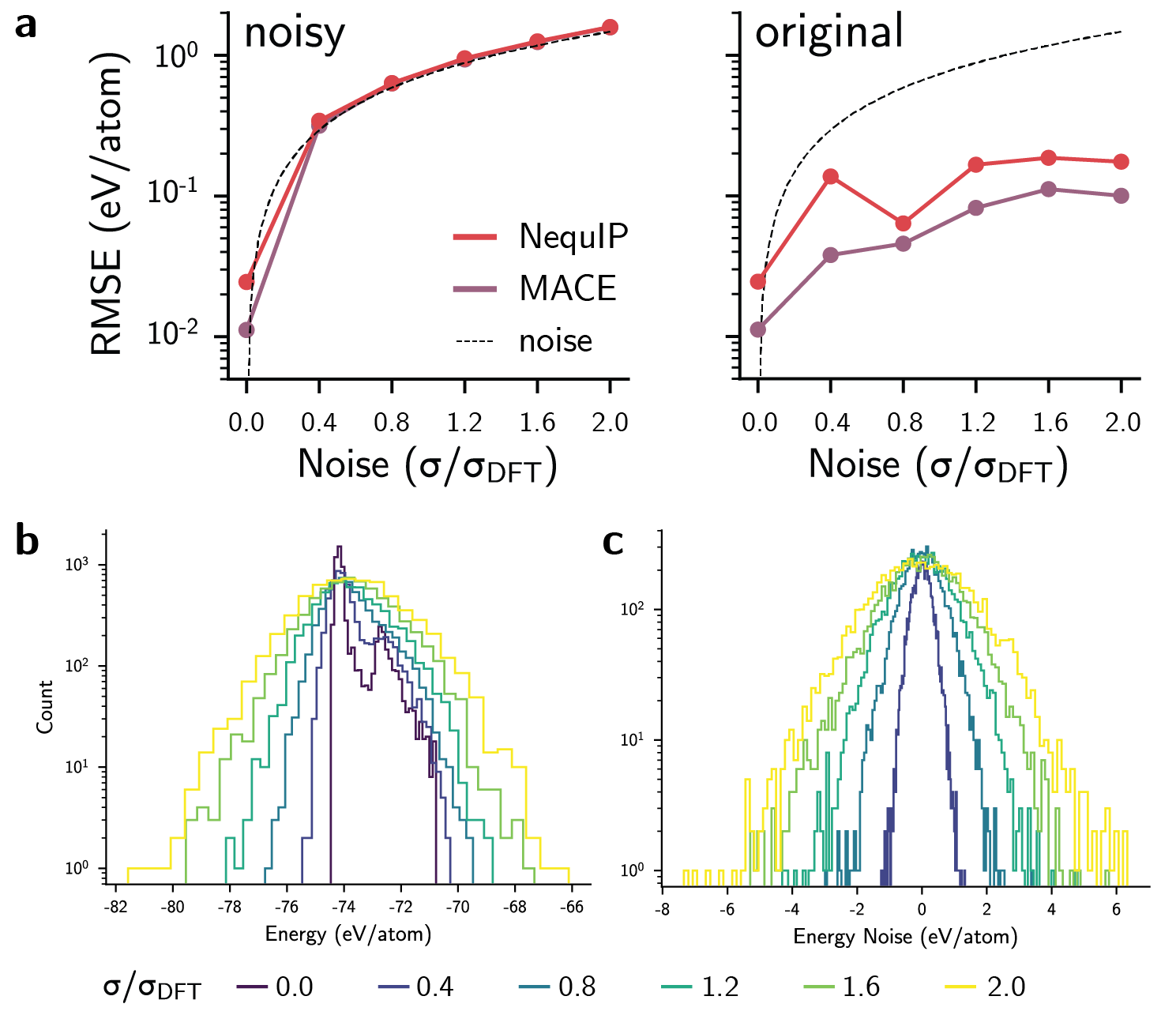}
    \caption{\textbf{a}, RMSE of energies (eV/atom) for models trained to highly noisy versions of the ANI-Al dataset, using noise magnitudes up to 20$\times$ those in Fig. \ref{fig:si:noisy_fits_ani}. The models trained on noisy data are then evaluated on their noisy training set (``noisy'' chart) or on the original, uncorrupted dataset (``original''). The black dashed lines correspond to the amount of noise that was added to the DFT energies in units of eV/atom. Distributions of \textbf{b}, energies and \textbf{c}, forces noises for this high-noise ANI-Al dataset.}
    \label{fig:si:ani_high_noise}
\end{figure}

\clearpage
\subsection{Distributions of model weights}
\begin{figure}[!htb]
    \centering
    \includegraphics[width=\linewidth]{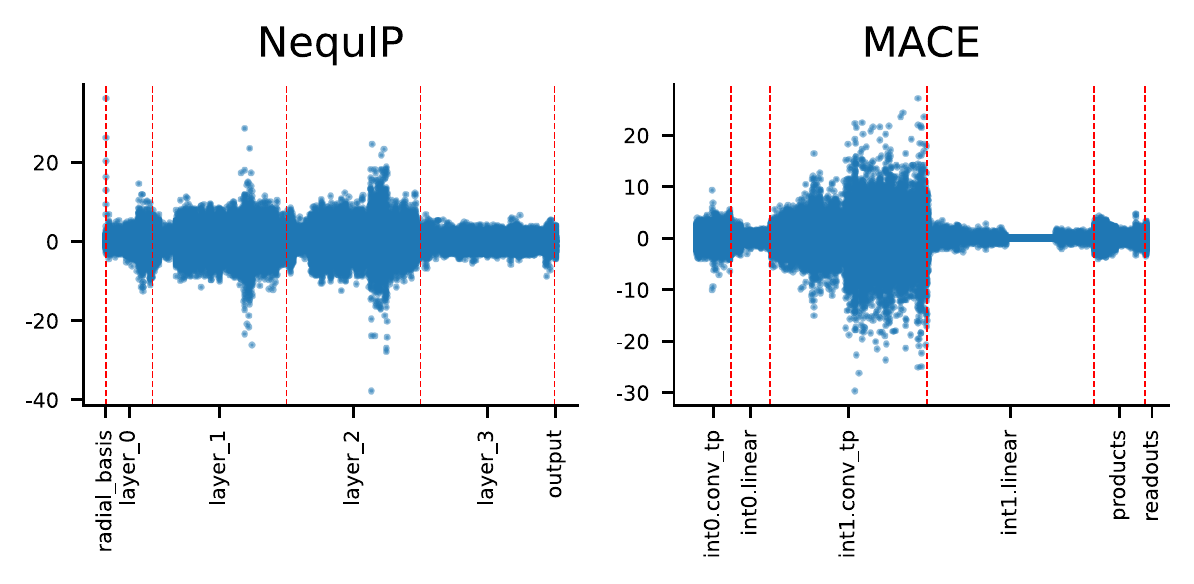}
        \caption{Plots of the parameters of the four models trained to the ANI-Al dataset. Dashed red lines are drawn to help highlight important parameter sets, where the labels indicate the (abbreviated) name of the parameter set. Parameter sets of note from each model include the Bessel function weights (NequIP) and convolutional tensor product layers (MACE).}
    \label{fig:si:model_parameters}
\end{figure}

\begin{figure*}[!htb]
    \centering
    \includegraphics[width=\smallfigwidth]{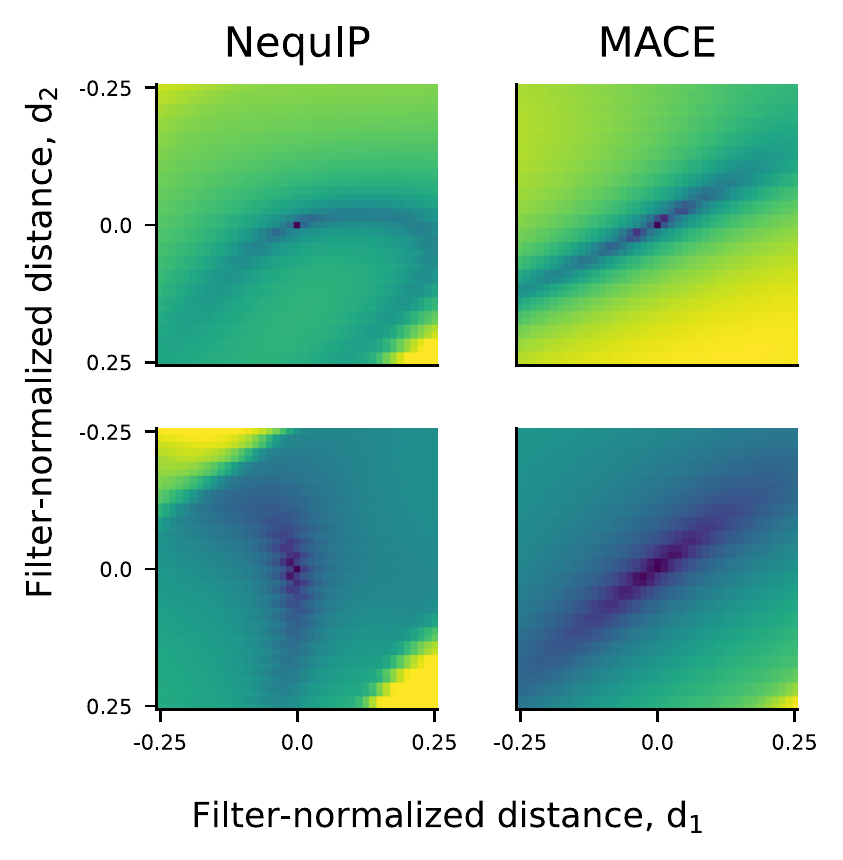}
    \caption{2D landscapes for the energy (top) and forces (bottom) loss of NNIP models on the ANI-Al dataset. The landscapes are plotted with respect to two normalized directions $d_1$ and $d_2$, as described in Sec. \ref{sec:methods:ll}. Colors correspond to the same values of losses across models, with brighter colors corresponding to higher losses.}
    \label{fig:si:2d_ll_aluminum}
\end{figure*}

\clearpage

\clearpage
\section{Supplementary Tables}

\begin{table}[h]
\centering
\caption{Major hyperparameters used to build/train the NequIP models in this work.}
\label{tab:si:nequip_hparams}
\begin{tabular}{lccccc}
\hline \hline
Model & Num. Layers & AMSGrad & EMA & Train. Bessel & Rescaled \\ 
 \hline no rescaling & 2 & \TrueMarker & \TrueMarker & \TrueMarker & \FalseMarker \\ 
 rescaling & 2 & \TrueMarker & \TrueMarker & \FalseMarker & \TrueMarker \\ 
 rescaling + bessel & 2 & \TrueMarker & \TrueMarker & \TrueMarker & \TrueMarker \\
 \hline
 2-layer, baseline & 2 & \FalseMarker & \FalseMarker & \TrueMarker & \TrueMarker \\ 
 2-layer, AMSGrad-only & 2 & \TrueMarker & \FalseMarker & \TrueMarker & \TrueMarker \\ 
 2-layer, EMA-only & 2 & \FalseMarker & \TrueMarker & \TrueMarker & \TrueMarker \\ 
 \hline
 5-layer, baseline & 5 & \FalseMarker & \FalseMarker & \FalseMarker & \TrueMarker \\ 
 5-layer, AMSGrad-only & 5 & \TrueMarker & \FalseMarker & \FalseMarker & \TrueMarker \\ 
 5-layer, EMA-only & 5 & \FalseMarker & \TrueMarker & \FalseMarker & \TrueMarker \\ 
 \hline
 2-layer & 2 & \TrueMarker & \TrueMarker & \FalseMarker & \TrueMarker \\ 
 3-layer & 3 & \TrueMarker & \TrueMarker & \FalseMarker & \TrueMarker \\ 
 4-layer & 4 & \TrueMarker & \TrueMarker & \FalseMarker & \TrueMarker \\ 
 5-layer & 5 & \TrueMarker & \TrueMarker & \FalseMarker & \TrueMarker \\ 
 \hline \hline \end{tabular}
\end{table}

\begin{table}[h]
\centering
\caption{Results of LLs entropy and MD simulation stability for NequIP models. The values of entropy of the energy loss ($S_e$) and forces loss ($S_f$) are computed with the parameters discussed in Sec. \ref{sec:methods:entropy}. The error bar of the time to failure is the standard deviation of all 30 trajectories performed for each model.}
\label{tab:si:nequip_entropy}
\begin{tabular}{lcccc}
\hline \hline
Model & $S_e$ & $S_f$ & $S$ & Time to failure (ps) \\ 
\hline
no rescaling & -1.53 & -0.67 & -0.84 & $1.65 \pm $0.38 \\ 
rescaling + Bessel & 0.33 & 1.78 & 1.49 & $2.34 \pm $1.27 \\ 
rescaling & 0.26 & 1.90 & 1.57 & $2.84 \pm $1.26 \\ 
\hline
2-layer, baseline & 0.13 & 1.98 & 1.61 & $2.04 \pm $0.76 \\ 
2-layer, AMSGrad-only & 0.35 & 1.78 & 1.50 & $2.24 \pm $1.14 \\ 
2-layer, EMA-only & -0.13 & 1.87 & 1.47 & $1.51 \pm $0.45 \\ 
\hline
5-layer, baseline & 2.02 & 2.13 & 2.11 & $2.97 \pm $1.54 \\ 
5-layer, AMSGrad-only & 2.29 & 2.35 & 2.34 & $3.62 \pm $1.65 \\ 
5-layer, EMA-only & 1.41 & 2.16 & 2.01 & $2.27 \pm $1.03 \\ 
\hline
2-layer & 0.33 & 1.80 & 1.51 & $1.96 \pm $0.59 \\ 
3-layer & 0.09 & 2.34 & 1.89 & $3.64 \pm $1.82 \\ 
4-layer & 0.19 & 2.48 & 2.02 & $3.67 \pm $1.64 \\ 
5-layer & 2.14 & 2.31 & 2.28 & $4.62 \pm $1.63 \\ 
\hline \hline
\end{tabular} \end{table}

\begin{table}[h]
\centering
\caption{Test RMSE of energies and forces in the 3BPA dataset for most NequIP models in this study. All models were trained on different dataset sizes ($N \in \{25, 125, 250, 500\}$) containing samples from simulations at 300 K. Then, models were tested against held-out samples from simulations at 300, 600, and 1200 K.}
\label{tab:si:nequip_rmse}
\begin{tabular}{|l|c|cccc|cccc|} \hline \hline
& & \multicolumn{4}{|c|}{Energy RMSE (meV/atom)} & \multicolumn{4}{|c|}{Forces RMSE (meV/\AA)} \\
\hline
Model & T & 25 & 125 & 250 & 500 & 25 & 125 & 250 & 500 \\ 
\hline
& 300 & 2.7 & 1.0 & 1.3 & 1.6 & 92.1 & 45.9 & 34.0 & 25.7 \\ 
2-layer, baseline & 600 & 3.3 & 1.7 & 1.7 & 1.8 & 140.8 & 86.5 & 68.2 & 54.4 \\ 
& 1200 & 5.8 & 3.8 & 3.7 & 3.7 & 248.0 & 181.6 & 161.8 & 146.3 \\ 
\hline
& 300 & 3.5 & 1.1 & 0.7 & 0.3 & 91.7 & 45.0 & 33.6 & 25.0 \\ 
2-layer, AMSGrad-only & 600 & 3.6 & 1.7 & 1.2 & 0.8 & 139.2 & 83.8 & 67.5 & 52.5 \\ 
& 1200 & 5.4 & 3.5 & 2.8 & 2.5 & 248.6 & 173.4 & 154.0 & 132.9 \\ 
\hline
& 300 & 2.7 & 1.0 & 0.5 & 3.7 & 93.2 & 45.6 & 34.0 & 25.4 \\ 
2-layer, EMA-only & 600 & 3.4 & 1.7 & 1.1 & 3.7 & 143.2 & 86.3 & 68.3 & 54.1 \\ 
& 1200 & 6.0 & 3.8 & 3.3 & 5.6 & 251.0 & 181.4 & 168.6 & 151.8 \\ 
\hline
& 300 & 2.0 & 2.8 & 8.3 & 9.4 & 88.7 & 30.0 & 20.4 & 16.1 \\ 
5-layer, baseline & 600 & 2.9 & 2.8 & 8.1 & 9.2 & 133.8 & 57.3 & 43.0 & 34.9 \\ 
& 1200 & 6.0 & 3.7 & 8.9 & 9.9 & 233.4 & 129.0 & 115.0 & 98.8 \\ 
\hline
& 300 & 2.1 & 0.5 & 1.5 & 0.3 & 87.8 & 28.6 & 19.4 & 14.4 \\ 
5-layer, AMSGrad-only & 600 & 3.1 & 1.0 & 1.6 & 0.6 & 134.1 & 55.0 & 40.8 & 31.2 \\ 
& 1200 & 6.6 & 2.1 & 2.8 & 1.9 & 239.6 & 121.2 & 103.3 & 85.2 \\ 
\hline
& 300 & 1.9 & 1.1 & 2.2 & 5.3 & 88.8 & 29.1 & 20.1 & 14.5 \\ 
5-layer, EMA-only & 600 & 2.9 & 1.4 & 2.3 & 5.3 & 133.6 & 56.5 & 44.0 & 33.8 \\ 
& 1200 & 5.8 & 2.8 & 4.2 & 3.7 & 232.5 & 132.7 & 121.9 & 101.3 \\ 
\hline
& 300 & 6.7 & 2.6 & 10.1 & 0.9 & 89.8 & 43.7 & 33.2 & 25.2 \\ 
2-layer & 600 & 7.9 & 2.8 & 10.1 & 1.2 & 136.8 & 82.0 & 65.7 & 52.6 \\ 
& 1200 & 10.4 & 5.5 & 11.9 & 3.0 & 243.6 & 175.3 & 153.6 & 138.9 \\ 
\hline
& 300 & 33.5 & 0.5 & 0.4 & 1.7 & 93.2 & 33.1 & 22.4 & 15.3 \\ 
3-layer & 600 & 34.9 & 1.1 & 0.8 & 1.7 & 139.0 & 61.5 & 45.8 & 34.4 \\ 
& 1200 & 37.8 & 2.5 & 1.9 & 2.6 & 242.2 & 131.1 & 112.2 & 92.8 \\ 
\hline
& 300 & 1.9 & 0.4 & 0.4 & 0.6 & 95.1 & 31.3 & 19.7 & 12.5 \\ 
4-layer & 600 & 2.9 & 1.0 & 0.8 & 0.8 & 143.9 & 58.9 & 41.3 & 29.7 \\ 
& 1200 & 6.5 & 2.3 & 1.7 & 1.7 & 258.9 & 126.5 & 101.9 & 85.0 \\ 
\hline
& 300 & 2.0 & 0.5 & 0.5 & 2.7 & 87.7 & 28.7 & 19.0 & 13.1 \\ 
5-layer & 600 & 2.9 & 1.0 & 0.8 & 2.7 & 133.2 & 55.0 & 40.5 & 30.5 \\ 
& 1200 & 6.1 & 2.1 & 2.1 & 3.6 & 235.8 & 121.5 & 104.6 & 84.2 \\ 
\hline \hline
\end{tabular} \end{table}

\begin{table}[h]
\centering
\caption{Major hyperparameters used to build/train the MACE models in this work. The model ``rescaling + Bessel'' is the only one containing trainable Bessel functions, which were introduced in the MACE code for this ablation study.}
\label{tab:si:mace_hparams}
\begin{tabular}{lclccccc}
\hline \hline
Model & $v$ & hidden\_irreps & $L$ & AMSGrad & EMA & SWA + WC & Rescaled \\ 
\hline
no rescaling & 2 & 256x0e + 256x1o + 256x2e & 3 & \TrueMarker & \TrueMarker & \TrueMarker & \FalseMarker \\ 
rescaling + opt & 2 & 256x0e + 256x1o + 256x2e & 3 & \TrueMarker & \TrueMarker & \TrueMarker & \TrueMarker \\ 
rescaling & 2 & 256x0e + 256x1o + 256x2e & 3 & \FalseMarker & \FalseMarker & \FalseMarker & \TrueMarker \\ 
rescaling + Bessel & 2 & 256x0e + 256x1o + 256x2e & 3 & \TrueMarker & \TrueMarker & \TrueMarker & \TrueMarker \\ 
\hline
$v=2, L=3$, none & 2 & 256x0e + 256x1o + 256x2e & 3 & \FalseMarker & \FalseMarker & \FalseMarker & \TrueMarker \\ 
$v=2, L=3$, SWA + WC & 2 & 256x0e + 256x1o + 256x2e & 3 & \FalseMarker & \FalseMarker & \TrueMarker & \TrueMarker \\ 
$v=2, L=3$, EMA & 2 & 256x0e + 256x1o + 256x2e & 3 & \FalseMarker & \TrueMarker & \FalseMarker & \TrueMarker \\ 
$v=2, L=3$, AMSGrad & 2 & 256x0e + 256x1o + 256x2e & 3 & \TrueMarker & \FalseMarker & \FalseMarker & \TrueMarker \\ 
\hline
$v=2, L=0$ & 2 & 256x0e & 0 & \TrueMarker & \TrueMarker & \TrueMarker & \TrueMarker \\ 
$v=2, L=1$ & 2 & 256x0e + 256x1o & 1 & \TrueMarker & \TrueMarker & \TrueMarker & \TrueMarker \\ 
$v=2, L=2$ & 2 & 256x0e + 256x1o + 256x2e & 2 & \TrueMarker & \TrueMarker & \TrueMarker & \TrueMarker \\ 
$v=2, L=3$ & 2 & 256x0e + 256x1o + 256x2e & 3 & \TrueMarker & \TrueMarker & \TrueMarker & \TrueMarker \\ 
\hline
$v=1, L=3$ & 2 & 256x0e + 256x1o + 256x2e & 3 & \TrueMarker & \TrueMarker & \TrueMarker & \TrueMarker \\ 
$v=2, L=3$ & 2 & 256x0e + 256x1o + 256x2e & 3 & \TrueMarker & \TrueMarker & \TrueMarker & \TrueMarker \\ 
$v=3, L=3$ & 2 & 256x0e + 256x1o + 256x2e & 3 & \TrueMarker & \TrueMarker & \TrueMarker & \TrueMarker \\ 
\hline \hline
\end{tabular}
\end{table}

\begin{table}[h]
\centering
\caption{Results of LLs entropy and MD simulation stability for MACE models. The values of entropy of the energy loss ($S_e$) and forces loss ($S_f$) are computed with the parameters discussed in Sec. \ref{sec:methods:entropy}. The error bar of the time to failure is the standard deviation of all 30 trajectories performed for each model.}
\label{tab:si:mace_entropy}
\begin{tabular}{lcccc}
\hline \hline
Model & $S_e$ & $S_f$ & $S$ & Time to failure (ps) \\ 
\hline
no rescaling & -0.31 & -0.05 & -0.11 & $0.39 \pm $0.00 \\ 
rescaling + opt & 0.63 & 2.48 & 2.11 & $2.53 \pm $1.40 \\ 
rescaling & 0.28 & 2.45 & 2.02 & $1.94 \pm $0.90 \\ 
rescaling + Bessel & 0.52 & 2.53 & 2.13 & $1.86 \pm $0.81 \\ 
\hline
$v=2, L=3$, none & 0.28 & 2.45 & 2.02 & $2.25 \pm $1.37 \\ 
$v=2, L=3$, SWA + WC & 0.55 & 2.56 & 2.16 & $2.68 \pm $1.30 \\ 
$v=2, L=3$, EMA & 0.08 & 2.45 & 1.97 & $2.13 \pm $0.90 \\ 
$v=2, L=3$, AMSGrad & 0.48 & 2.48 & 2.08 & $2.71 \pm $1.51 \\ 
\hline
$v=2, L=0$ & 0.92 & 1.93 & 1.73 & $0.39 \pm $0.00 \\ 
$v=2, L=1$ & 1.15 & 2.74 & 2.42 & $0.89 \pm $0.10 \\ 
$v=2, L=2$ & 0.42 & 2.47 & 2.06 & $3.42 \pm $1.97 \\ 
$v=2, L=3$ & 0.47 & 2.44 & 2.04 & $1.72 \pm $0.91 \\ 
\hline
$v=1, L=3$ & 2.04 & 2.17 & 2.14 & $1.92 \pm $0.84 \\ 
$v=2, L=3$ & 0.47 & 2.44 & 2.04 & $1.72 \pm $0.91 \\ 
$v=3, L=3$ & 0.63 & 2.64 & 2.24 & $2.46 \pm $1.12 \\ 
\hline \hline
\end{tabular} \end{table}

\begin{table}[h]
\centering
\caption{Test RMSE of energies and forces in the 3BPA dataset for most MACE models in this study. All models were trained on different dataset sizes ($N \in \{25, 125, 250, 500\}$) containing samples from simulations at 300 K. Then, models were tested against held-out samples from simulations at 300, 600, and 1200 K. The entry $v = 2, L = 3$ is duplicated in this table to facilitate the visualization of the trends.}
\label{tab:si:mace_rmse}
\begin{tabular}{|l|c|cccc|cccc|}
\hline \hline
& & \multicolumn{4}{|c|}{Energy RMSE (meV/atom)} & \multicolumn{4}{|c|}{Forces RMSE (meV/\AA)} \\
\hline
Model & T (K) & 25 & 125 & 250 & 500 & 25 & 125 & 250 & 500 \\
\hline
 & 300 & 3.3 & 0.7 & 0.3 & 0.1 & 65.3 & 23.9 & 17.3 & 12.0 \\ 
$v=2, L=3$, none & 600 & 3.6 & 1.0 & 0.6 & 0.4 & 102.4 & 47.8 & 39.9 & 31.0 \\ 
 & 1200 & 5.9 & 1.9 & 1.9 & 1.6 & 203.9 & 127.2 & 121.9 & 107.2 \\ 
\hline
 & 300 & 1.5 & 0.3 & 0.2 & 0.2 & 71.5 & 26.2 & 18.3 & 12.2 \\ 
$v=2, L=3$, SWA + WC & 600 & 1.8 & 0.8 & 0.6 & 0.5 & 116.1 & 50.2 & 41.3 & 31.5 \\ 
 & 1200 & 5.6 & 2.2 & 2.1 & 1.9 & 243.5 & 128.4 & 125.8 & 109.8 \\ 
\hline
 & 300 & 2.2 & 0.3 & 0.2 & 0.7 & 66.6 & 24.4 & 17.5 & 12.0 \\ 
$v=2, L=3$, EMA & 600 & 2.5 & 0.8 & 0.6 & 0.8 & 104.4 & 49.6 & 40.2 & 31.0 \\ 
 & 1200 & 5.0 & 2.3 & 1.9 & 2.1 & 208.8 & 136.6 & 121.3 & 109.1 \\ 
\hline
 & 300 & 1.9 & 0.4 & 0.2 & 0.1 & 64.3 & 23.6 & 16.1 & 10.6 \\ 
$v=2, L=3$, AMSGrad & 600 & 2.0 & 0.8 & 0.5 & 0.4 & 103.4 & 46.6 & 35.5 & 26.7 \\ 
 & 1200 & 4.1 & 1.9 & 1.6 & 1.4 & 207.9 & 117.1 & 98.2 & 87.3 \\ 
\hline
 & 300 & 4.7 & 1.3 & 0.8 & 0.6 & 194.5 & 90.6 & 61.4 & 47.5 \\ 
$v = 2, L = 0$ & 600 & 11.1 & 4.5 & 2.3 & 2.0 & 311.3 & 179.6 & 130.1 & 108.4 \\ 
 & 1200 & 14.2 & 9.0 & 6.3 & 5.3 & 491.6 & 353.8 & 320.6 & 288.1 \\ 
\hline
 & 300 & 3.0 & 0.6 & 0.4 & 0.2 & 125.6 & 42.7 & 29.0 & 19.4 \\ 
$v = 2, L = 1$ & 600 & 4.1 & 1.5 & 1.1 & 0.7 & 188.7 & 84.8 & 63.4 & 46.9 \\ 
 & 1200 & 11.0 & 5.0 & 3.9 & 2.8 & 346.2 & 196.5 & 163.8 & 138.7 \\ 
\hline
 & 300 & 1.3 & 0.3 & 0.2 & 0.1 & 70.3 & 27.0 & 18.3 & 12.5 \\ 
$v = 2, L = 2$ & 600 & 2.1 & 0.8 & 0.6 & 0.5 & 115.2 & 53.4 & 41.8 & 31.1 \\ 
 & 1200 & 4.7 & 2.1 & 1.9 & 1.6 & 218.6 & 127.9 & 115.7 & 97.1 \\ 
\hline
 & 300 & 1.4 & 0.3 & 0.2 & 0.1 & 66.6 & 24.4 & 16.9 & 11.0 \\ 
$v = 2, L = 3$ & 600 & 1.8 & 0.8 & 0.6 & 0.4 & 106.8 & 48.5 & 37.5 & 27.6 \\ 
 & 1200 & 4.6 & 2.2 & 1.9 & 1.6 & 218.3 & 125.7 & 107.1 & 93.3 \\ 
\hline
 & 300 & 1.9 & 0.7 & 0.6 & 0.5 & 126.6 & 55.4 & 42.9 & 34.6 \\ 
$v = 1, L = 3$ & 600 & 3.7 & 1.9 & 1.4 & 1.2 & 188.2 & 102.9 & 85.3 & 73.7 \\ 
 & 1200 & 8.0 & 4.7 & 4.1 & 3.6 & 330.1 & 216.0 & 200.8 & 189.4 \\ 
\hline
 & 300 & 1.4 & 0.3 & 0.2 & 0.1 & 66.6 & 24.4 & 16.9 & 11.0 \\ 
$v = 2, L = 3$ & 600 & 1.8 & 0.8 & 0.6 & 0.4 & 106.8 & 48.5 & 37.5 & 27.6 \\ 
 & 1200 & 4.6 & 2.2 & 1.9 & 1.6 & 218.3 & 125.7 & 107.1 & 93.3 \\ 
\hline
 & 300 & 1.2 & 0.3 & 0.2 & 0.1 & 66.0 & 23.5 & 15.8 & 10.6 \\ 
$v = 3, L = 3$ & 600 & 1.8 & 0.8 & 0.6 & 0.5 & 108.7 & 48.9 & 36.9 & 28.0 \\ 
 & 1200 & 6.1 & 3.1 & 2.6 & 2.2 & 226.5 & 137.9 & 115.5 & 101.1 \\ 
\hline \hline
\end{tabular} \end{table}

\end{document}